# Prognosis of Multivariate Battery State of Performance and Health via Transformers

Noah H. Paulson[1,*], Joseph J. Kubal[2], Susan J. Babinec[3]

[1]Data Science and Learning Division, Argonne National Laboratory, Argonne, IL 60439
[2]Chemical Sciences and Engineering Division, Argonne National Laboratory, Argonne, IL 60439
[3]Argonne Collaborative Center for Energy Storage Science (ACCESS), Argonne National Laboratory, Argonne, IL 60439

## Abstract

Batteries are an essential component in a deeply decarbonized future. Understanding battery performance and "useful life" as a function of design and use is of paramount importance to accelerating adoption. Historically, battery state of health (SOH) was summarized by a single parameter, the fraction of a battery's capacity relative to its initial state. A more useful approach, however, is a comprehensive characterization of its state and complexities, using an interrelated set of descriptors including capacity, energy, ionic and electronic impedances, open circuit voltages, and microstructure metrics. Indeed, predicting across an extensive suite of properties as a function of battery use is a 'holy grail' of battery science; it can provide unprecedented insights toward the design of better batteries with reduced experimental effort, and de-risking energy storage investments that are necessary to meet $CO_2$ reduction targets. In this work, we present a first step in that direction via deep transformer networks for the prediction of 28 battery state of health descriptors using two cycling datasets representing six lithium-ion cathode chemistries (LFP, NMC111, NMC532, NMC622, HE5050, and 5Vspinel), multiple electrolyte/anode compositions, and different charge-discharge scenarios. The accuracy of these predictions versus battery life (with an unprecedented mean absolute error of 19 cycles in predicting end of life for an LFP fast-charging dataset) illustrates the promise of deep learning towards providing deeper understanding and control of battery health.

# 1. Introduction

The urgency of addressing climate change dictates a pace of innovation which is unfamiliar by historic norms. The transition of today's energy infrastructure to one which is fully renewables-driven depends on the development of diverse energy storage technologies that are deployed at a massive scale. The financial demands of this transformation are commensurate with the critical need – exceptionally large investments are required across the technology life spectrum from research, which requires years in the lab, to deployments, where capital outlays for large-scale renewables systems are enormous, even for mature technologies. Adding to these already significant demands is ubiquitous, path-dependent device degradation, which further intensifies risk, especially considering that these markets are evolving while energy storage technologies are under development. Thus, it is of paramount importance to attenuate these fundamental risks by improving our predictive capabilities across diverse performance and health characteristics throughout the device lifetime. In the laboratory, rapid and comprehensive diagnosis of device health and prognosis of future health and performance will be critical to accelerate screening and optimization of new technologies, while in the field, this same capability is critical to optimal design and operation of field investments that must perform over a 20-year lifetime.

Comprehensive battery health diagnosis and prognosis of today's leading energy storage technology – Li-ion – remains challenging despite its critical role in the renewable ecosystem. Determination of Li-ion battery lifetime and degradation modes via experiment is expensive and requires months to years; physics-based models provide a promising alternative but are challenging to develop and are computationally expensive to run. Data-driven methods have emerged in recent years for the diagnosis and prognosis of Li-ion battery state of charge (SOC) and state of health (SOH) – often defined as both available capacity and internal resistance. Earlier works focused on predicting the battery SOC [1] and SOH [2–7] via statistical filtering methods, traditional machine learning (ML), and deep learning (DL). The introduction of an open-source cycling dataset for the fast-charging of 124 commercial lithium-iron-phosphate (LFP) cells by Severson *et al.* in 2019 [8] transitioned the focus from SOC estimation and capacity prognosis of single cells [9] to prediction of the remaining useful life (RUL), defined as the number of cycles to failure [10–13]. Using the same datasets, a second generation of papers demonstrated prognosis of capacity and resistance with the new capability to predict the roll-over of these quantities using between 1 and 100 preliminary cycles to inform these predictions: in other words - the context [14,15], sometimes with quantified uncertainty [16–21]. Furthermore, sequence-to-sequence auto-encoders have emerged as a dominant DL prognosis strategy for battery state of health [21–23].

Over the past several years, the battery community has begun to recognize that the prediction of battery capacity and internal resistance is insufficient to accurately portray a complex battery state of performance and health, not only for Li-ion but in fact for all storage technologies. Batteries degrade in different ways and different rates depending on their chemistry, design, and the way that they are used. For example, Li-ion batteries charged at high rates may suffer from Li-plating, which leads to a typically unpredictable, future rapid decrease in capacity. In contrast, cells that are cycled at low rates but used at high temperatures sometimes form thick solid electrolyte interfaces that increase resistances, leading to power deficits. In 2017, Birkl *et al.* advanced the state-of-the-art by demonstrating that Li-ion battery degradation could be categorized as loss of lithium inventory (LLI), loss of active material of the negative electrode ($LAM_{NE}$), and loss of active material of the positive electrode ($LAM_{PE}$) by producing and cycling coin cells with specific levels of degradation [24]. This approach employed a vector of performance and health quantities that offered insights beyond the usual capacity and internal resistance values. Dubarry *et al.* constructed a database of simulated degradation profiles for Li-ion batteries including LLI, $LAM_{NE}$, and $LAM_{PE}$ [25], and Chen *et al.* demonstrated ML diagnosis of these degradation modes in two NMC cathode chemistries, and early prognosis of the end-of-life $LAM_{PE}$ [26]. Kim et al. employed deep learning informed by simulation and limited experiment to predict LLI, LAM, and end of life capacity for a small experimental dataset [27]. Recent work by Aitio *et al.* demonstrated Gaussian process regression in describing equivalent circuit model (ECM) parameters as a function of use conditions and time, hinting at prognosis of these parameters as well [28]. While these advances are promising, it has become clear that an even more comprehensive understanding of the battery health and performance will motivate urgently needed commercial and scientific investments - *we refer to this vector as the multivariate state of performance and health (SoPaH)*.

In this work, we propose an advancement to the state-of-the-art by introducing and demonstrating DL prognosis of multivariate battery SoPaH for a variety of cathode, anode, and electrolyte chemistries, and two use scenarios. This is an important step towards the capabilities required to account for a rapidly changing grid that will require different uses for batteries across their lifetimes. In contrast to existing work that only predicts degradation trajectories for capacity and internal resistance, we offer predictions on capacity, energy, round-trip efficiencies, internal resistance, and ECM extracted open circuit voltage (OCV) and resistance (R), as a function of SOC. Towards this new prognosis capability, we employed a multivariate transformer architecture introduced by Grigsby *et al.* [29] to predict these quantities as a function of the start cycle and forecast envelope, using only 1 to 64 preliminary cycles. *In other words, the model can predict long-duration SoPaH trends regardless of previous cycling history.* Beyond prediction of the usual quantities (capacity, energy, and resistance), ECM parameter prognosis enables prediction of the entire suite of battery responses with cycling. Through multivariate prognosis of the SoPaH for a variety of chemistries and use strategies we motivate the extension to a more comprehensive SoPaH that includes degradation modes, levels, and internal structure (either simulated or characterized via advanced experimental techniques). That is, this approach enables use of alternative electrochemical methods to inform the SoPaH diagnosis and prognosis, such as electrochemical impedance spectroscopy (EIS) and hybrid pulse power characterization (HPPC) and of materials characterization methods including X-ray, acoustic interrogation, and nuclear magnetic resonance. Accurate prognosis of these degradation and materials characteristics will be a transformative capability as it will allow for accelerated design of batteries for specific use cases and will dramatically de-risk the investments in electrochemical energy storage required to meet $CO_2$ reduction targets over the next decade.

The remainder of this work is organized as follows. Section 2 presents the datasets employed for the two case studies, introduces the multivariate transformer architecture, and details model configuration and training. Section 3 shows DL model results including SoPaH prognosis for example cells, end of life (EOL) prediction accuracy, and SoPaH prediction accuracy versus starting cycle and prognosis cycle. Section 4 provides a discussion of the results and Section 5 provides final conclusions.

# 2. Data and Methods
## 2a. Dataset Description
In this work, we employ two datasets that are available in the open literature. The first single-cathode chemistry dataset compares various fast charging protocols, and its dominant Li-plating degradation mechanism, with discharge as 4C constant current (CC) and constant voltage (CV). The second dataset investigates regular 100% depth of discharge (DoD) cycling at low charge and discharge rates across diverse cathode chemistries, covering a mix of degradation mechanisms. Using these contrasting designs, we hope to demonstrate the efficacy of the multivariate prognosis approach for the fast-charging scenario and across battery chemistries at low charge and discharge rates. These datasets do not address realistic use scenarios in that the cycles are regular and without variation – this will be left for future work.

The first dataset consisted of 178 graphite/LiFePO$_4$ (Gr/LFP) cylindrical cells produced by A123 Systems, cycled under fast-charging protocols as described in Severson *et al.* [8] and Attia *et al.* [30]. The cells had a nominal capacity and voltage of 1.1Ah and 3.3V, respectively, and were tested at a controlled ambient temperature of 30ºC. Individual cells were discharged at 4C until a lower cut-off voltage of 2.0V, followed by a CV discharge until a current cut-off of C/50 and C/20, in [8] and [30], respectively. Cells were charged in one or two high-rate CC steps up to 80% SOC, followed by a 1C CC charge up to 3.6V, followed by a CV charge until a current cut-of C/50 and C/20, in [8] and [30], respectively.

The second dataset consisted of 300 Li-ion pouch cells corresponding to six different cathode chemistries (NMC111, NMC532, NMC622, NMC811, HE5050, and 5Vspinel), with several electrolyte and anode chemistries. Cell capacities were between 0.2 to >3 Ah. The cells were manufactured at the Argonne Cell Analysis, Modeling, and Prototyping (CAMP) facility using pilot-scale industrial equipment with stringent fabrication protocols. Charge and discharge segments were largely limited to 1C or slower. As in our previous work [31], only cycles with regular CC (and sometimes CV) were used, excluding pauses and HPPC tests.

In contrast to our prior work, where hundreds of physics-informed features were devised and evaluated to enable the prediction of battery EOL as defined by capacity, but which did not necessarily have any diagnostic power beyond that, here, the aim is to predict the evolution of key quantities (the SoPaH). In other words, the features and SoPaH are one in the same. For example, in applications where power is a critical variable, it will be useful to predict resistances versus use. This is a significant difference to prior efforts in two ways: (1) only features that provide direct physical or practical insights are selected and (2) it provides the ability to predict the evolution of those critical features with battery use. Many cycling datasets available today have limitations; for example, it is not currently possible to estimate open circuit voltage as a function of SOC. Instead, we relied on analysis of the available current and voltage versus time information in each cycle to extract the maximally useful SoPaH vector. Table 1 lists the SoPaH features of interest that could be extracted from each dataset. While an IR drop for the fast-charging data set was collected, the singular IR interruption occurred at variable SOC between the cells. This quantity was also excessively noisy and thus an ECM SOC-dependent resistance was used instead. ECMs were fit to the raw voltage response via simple IR circuit, where OCV and R were parameterized as a function of the %SOC through polynomial fits such that OCV was a monotonically increasing function [32]. These functions were then sampled every 10% SOC between 10% and 90% SOC to extract the ECM OCV and R features. Pauses between charging and discharging segments in the CAMP dataset enabled the extraction of the resistances associated with the instantaneous IR-drop at four different locations in the cycle. The minimum of these four resistance values was selected as the ohmic resistance value for that cycle to filter out erroneously large values caused by small timesteps.

Table 1: SoPaH vector components are listed for the fast-charging [8,30] and CAMP [33] datasets. The included X %SOC levels are [10, 20, …, 90].

| Feature of Potential Interest | Fast-charge | CAMP |
|---|---|---|
| Discharge capacity (Ah) | X | X |
| Discharge energy (Wh) | X | X |
| Discharge average voltage (V) | X | |
| Coulombic efficiency (%) | X | X |
| Energy efficiency (%) | X | X |
| Resistance IR-drop | | X |
| OCV at X% SOC from ECM (V) | X | |
| R at X% SOC from ECM (V) | X | |

It follows that AI trained with noisy or unreliable data leads to unreliable predictions. It was no different in this study, where unprocessed data was found to be problematic for some cells. Specifically, the time series data in the CAMP dataset was inherently noisier than the fast-charging data due to various irregular cycles and the use of experimental instead of commercial cells. For this reason, a filtering and segmenting strategy was developed to identify cycling regions with lower noise. First, the raw time-series were filtered using a custom median filter approach designed to reduce undesirable edge effects. Specifically, kernel sizes were reduced from 81 to 41 to 21 to 11 to 5 cycles depending on the distance from the first and last cycles. Next, cycles where the change in at least one of the filtered feature values exceeded some threshold (0.01 for capacity and energy, 0.007 for the efficiencies, and 0.02 for the resistance) were identified. Valid cycling segments were extracted from cycling regions where the number of cycles between "jump" cycles was at least 50, and the mean absolute percent error (MAPE) between the filtered segments and the unfiltered ones was less than some threshold (1% for capacity, energy, and the efficiencies, and 2% for the resistance). The values of these thresholds were determined in the training dataset by balancing the length and number of valid segments versus the noisiness of the time-series. No valid cycling segments were identified for cells with the NMC811 cathode chemistry. The same approach was applied for the fast-charge dataset but with less stringent thresholds. Some metrics, including the ECM-derived features, were especially noisy in the CAMP data, and were therefore not included. Further details can be found in our prior work [31], the original CAMP dataset metadata [33], and this work's codebase [34].

One barrier to working with deep learning models is the requirement for large datasets. In this work, we augmented the dataset by sub-sampling each segment. In other words, for a 1000 cycle segment using only 1 cycle of context information, 999 data points could be extracted. We acknowledge that data points sampled from the same segment are highly correlated, but without the addition of synthetic data from physics-based models, this was an effective alternative strategy. This approach also allowed for the examination of predictive performance versus starting cycle, an important capability for second-life applications.

## 2b. Multivariate Transformer Architecture

The transformer has arguably been the most important deep learning architecture in the field of artificial intelligence in the past decade [35]. In recent years, it has dominated the natural language processing space, culminating in tools, like Chat-GPT4, with exceptional performance in tasks including language translation, summarization, coding, and more. The transformer is a sequence-to-sequence deep learning architecture that leverages the self-attention mechanism to provide relational context for each token (i.e., element) of a given input sequence. Specifically, attention identifies tokens that are most relevant to each other token of interest (or query token). In contrast to recurrent neural network (RNN) architectures, which consider context for each token propagated forward or backwards through a series of processing units, the transformer takes the input as a whole, processing all elements in the sequence simultaneously. This ensures that all available information is considered across the input sequence, avoiding the vanishing gradient problem common to RNNs encountered when propagating learnings in a temporal manner.

In this work, we employ a modification of the transformer architecture, specifically for spatiotemporal forecasting (herein called the spacetimeformer), developed by Grigsby *et al.* [29]. Where the standard transformer can only take one-dimensional sequences as inputs and outputs, the spacetimeformer works with two-dimensional arrays that represent collections of correlated sequences. Use of the self-attention mechanism is expanded in-kind, where it provides relational context both in the temporal dimension and between sequences. The result is an algorithm that identifies and exploits correlations among time-series collected for the same system at the same time intervals. In other words, we employ the spacetimeformer architecture because it has exceptional performance for the prediction of the evolution of correlated time-series – in comparison to approaches that do not fully consider these correlations (e.g., traditional RNNs and transformers). Figure 1, reproduced from the original publication, compares the attention mechanisms employed in a univariate time-series transformer, a transformer employing spatial graphs within a given time-step, and the spatio-temporal attention mechanism employed in the spacetimeformer architecture. The shade of the lines connecting units of information in sub-figures b, c, and d illustrates the strength of the attention between temporal and spatial information. Note that in subfigure d, the attention is shared across spatial and temporal dimensions, and that the strength of the attention is not necessarily correlated with the spatial and temporal distances from the query token. The most important change to the algorithm was translating the typical sinusoidal features designed to give temporal context for minutes, hours, days, months, and years, to one, ten, and one-hundred cycles.

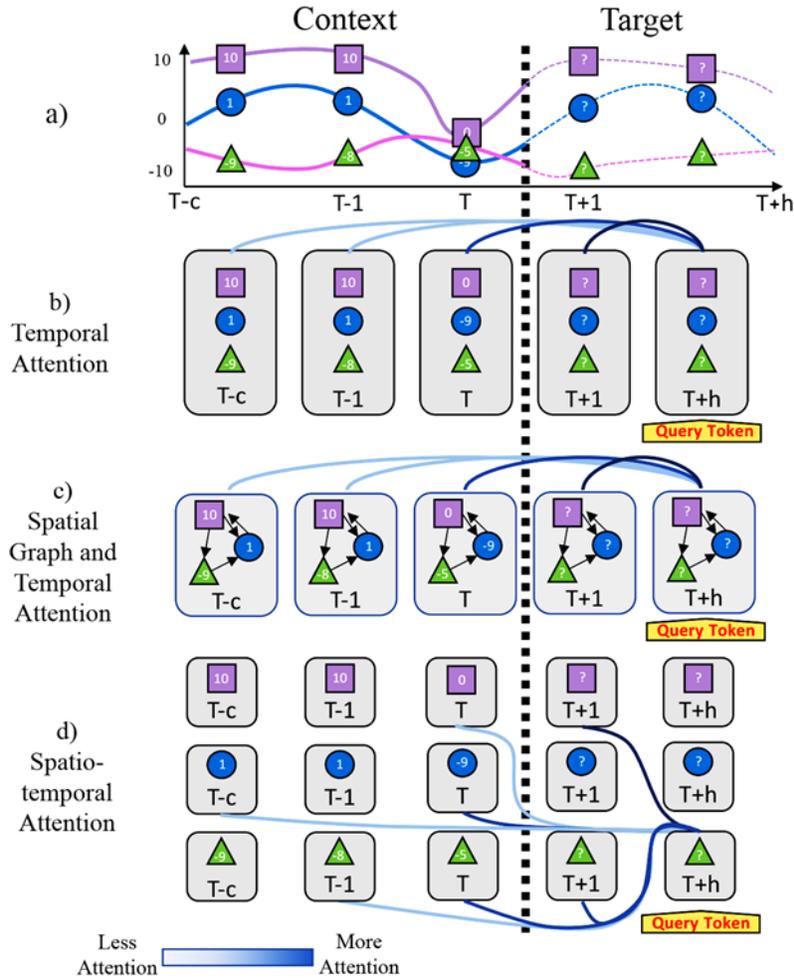

Figure 1: This figure, reproduced from Grigsby *et al.* [29], shows a) the form of the input data where the x-axis is time and the y axis shows feature values, b) the standard temporal attention mechanism exhibited in the standard transformer architecture, c) spatial graph with temporal attention, and d) spatio-temporal attention specific to the spacetimeformer architecture.

## 2c. Computational Approach

Training runs were performed on single A100 GPUs with 40GB memory on Argonne's Swing computing cluster. The Weights and Biases (W&B) framework was used to coordinate training runs [36]. W&B was configured to automatically collect and save information including the metadata associated with each run (e.g., date, wall-time, architecture details, etc.), use details of the computational resources, loss and error metrics, and all validation and test predictions. Furthermore, the W&B "sweeps" capability was employed to execute exploratory designs of experiments (DoE). Acceptable values of each hyperparameter were specified along with the search strategy. In this work, a Bayesian optimization strategy was selected with Hyperband [37] for early termination of runs with poor loss values compared to more successful runs. The default network architecture and hyperparameter configuration is listed in the SI.

# 3. Results

## 3.1 Hyperparameter Tuning

The most critical hyperparameter for the application of the model SoPaH diagnosis and prognosis is the number of cycles used as context. It is desirable to make the most accurate prediction with the fewest number of preliminary cycles, as this decreases the experimental costs of the prognosis effort. We investigated the relationship between the number of context cycles (exponentially increasing between 1 and 64) and validation metrics including mean absolute error (MAE) and mean squared error (MSE) through a series of training runs with the CAMP dataset. Bayesian inference using constant and linear models for the MSE and MAE on the validation set versus the number of context cycles revealed Bayes factors [38] of nearly 20 and of 1808 for MSE and MAE, respectively, for constant models over linear models. This indicated a strong Bayesian degree of preference for a constant model for MSE and a decisive preference based on MAE. These conclusions are mirrored in Figure 2, where the 95% credible interval for the best fit lines for the MSE and MAE losses are wide and represent the possibility of both positive and negative slopes. This preference for the constant model is likely due to the large degree of scatter in the loss metrics, i.e., the simpler model better describes the data. This outcome led us to use only one cycle as context for the remainder of this work. *Furthermore, it indicated that repeated training runs for a given hyperparameter configuration may be necessary to achieve an acceptable validation score.*

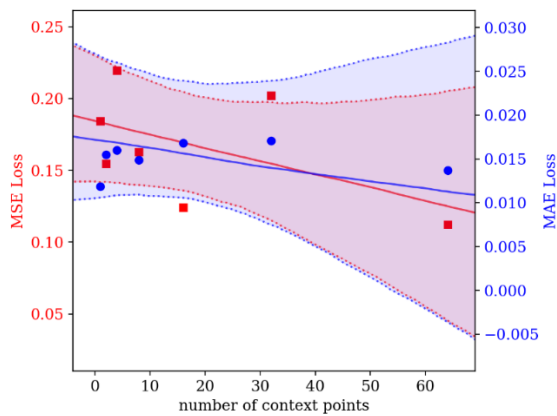

Figure 2: MSE and MAE losses for the CAMP dataset are plotted versus the number of cycles used as context. Ninety-five percent credible intervals are provided for the best fit lines for each loss metric. Statistical analysis led us to use only one cycle for context and to conclude that repeated training runs may be necessary for acceptable validation score.

## 3.2 SoPaH Prognosis

Figure 3 and Figure 4 plot representative experimental and predicted health and performance metric projections with cycling for cells from the fast-charging and CAMP datasets, respectively. The single preliminary cycle used by the AI model as context for the time-series forecasts is plotted in blue. Figure 3 is representative of the fast-charging-trained model predictions with high accuracy across all features. *Remarkably, AI accurately captures roll-over in all features, including complicated details of the ECM resistance and OCV at high SOC. It is unclear how (and why) the models are capturing these physical relationships as no physical principles are embedded, but these observations are consistent across AI for other scientific topics where deep learning captured physical insights.* We hypothesize that the reason for this exceptional performance is the use of both spatial and temporal attention in the transformer architecture. This allows the AI to leverage correlations across time series features to improve predictive accuracy for all. Figure 4 is representative of the CAMP-trained model predictions; prognosis accuracy is weaker, and the projected trend is offset from the context values. Hypotheses for this weaker performance are provided in the discussion section.

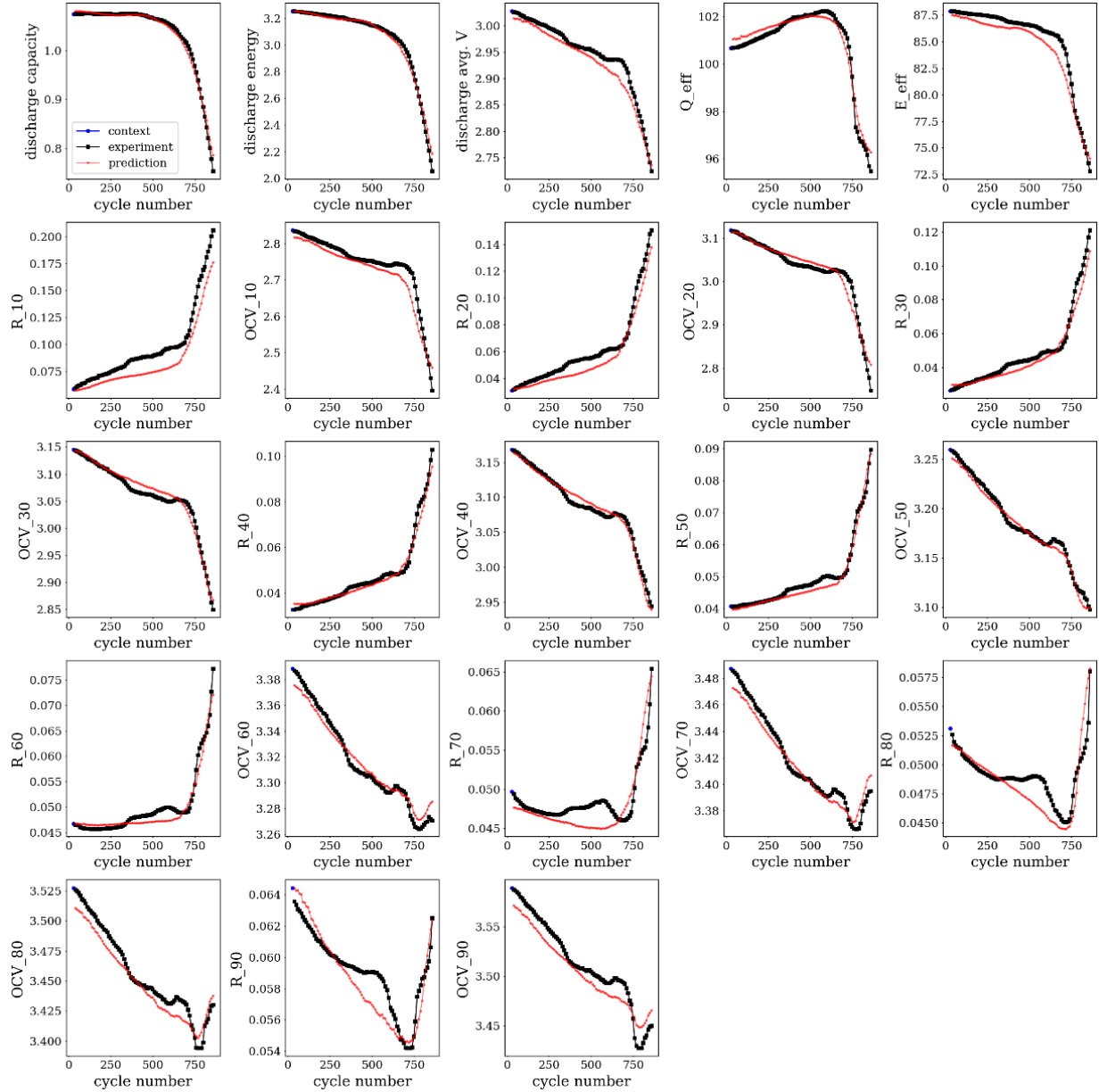

Figure 3: SoPaH prognosis versus cycling is plotted for fast-charged cell CH17 from Batch 1 of the Severson dataset from cycle 37 through 866. There is only a single context cycle and for this reason it may be difficult to distinguish from the remaining cycles in the figure. The first subplot (for discharge capacity) mirrors the current state of the art for SOH prognosis.

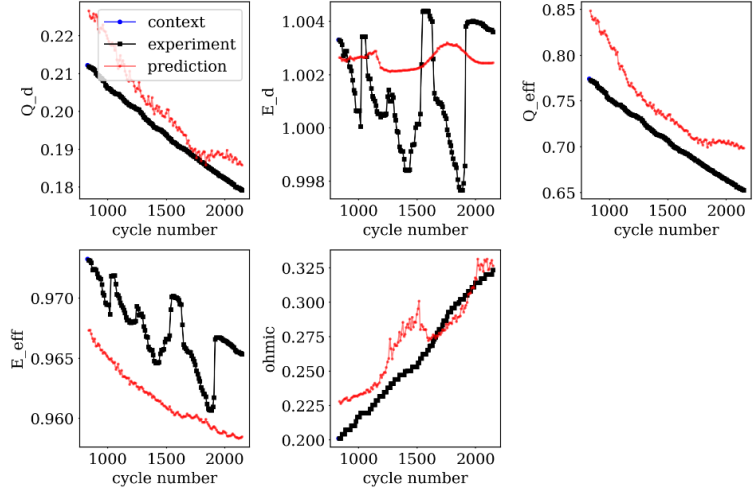

Figure 4: SoPaH prognosis versus cycling is plotted for the 3rd 5VSpinel cell in Batch B26Y, from cycle 837 to cycle 2154, across all SoPaH metrics.

Table 2 and Table 3 provide error metrics across the fast-charge and CAMP datasets, respectively, for all SoPaH metrics employed in the study. These error metrics were calculated on a cycle-by-cycle basis across all cycles and all cycling segments included in the test datasets. As such, they summarize general prognosis accuracy, not any traditional EOL metric (e.g., cycles until 80% of the original capacity is reached). As expected, dataset-wide performance is stronger for the fast-charge dataset than for the CAMP dataset, likely due to its inherently lower noise. MAEs are less than 2% for capacity and energy, less than 1% for the round-trip efficiencies and OCVs, and less than 8% for the resistances. The discrepancy in the prediction error between the ECM OCVs and resistances is notable, but the reason is unclear. In contrast, across the CAMP dataset, the prediction MAE exceeds 7% for capacity and coulombic efficiency and 30% for the ohmic resistance. Predictive performance for the energy and its round-trip efficiency are stronger, with MAEs of 0.1 and 0.6%, respectively. This may be a result of the relatively flat trend for these quantities.

Table 2: Error metrics, including mean absolute error (MAE), root mean squared error (RMSE), mean absolute percent error (MAPE), and root mean squared percent error (RMSPE), are provided for the fast-charge test dataset SoPaH metrics.

| Metric | MAE | RMSE | MAPE | RMSPE |
|---|---|---|---|---|
| discharge capacity | 0.014 +/- 0.011 | 0.018 +/- 0.013 | 1.49 +/- 1.18 | 1.28 +/- 1.08 |
| discharge energy | 0.049 +/- 0.037 | 0.059 +/- 0.042 | 1.72 +/- 1.34 | 1.46 +/- 1.24 |
| discharge avg. V | 0.012 +/- 0.009 | 0.013 +/- 0.01 | 0.401 +/- 0.307 | 0.347 +/- 0.325 |
| Q_eff | 0.207 +/- 0.221 | 0.252 +/- 0.278 | 0.206 +/- 0.222 | 0.131 +/- 0.167 |
| E_eff | 0.567 +/- 0.455 | 0.658 +/- 0.5 | 0.679 +/- 0.551 | 0.591 +/- 0.571 |
| R_10 | 0.007 +/- 0.006 | 0.008 +/- 0.006 | 7.66 +/- 6.39 | 7.04 +/- 6.79 |
| OCV_10 | 0.015 +/- 0.012 | 0.019 +/- 0.014 | 0.577 +/- 0.485 | 0.473 +/- 0.454 |
| R_20 | 0.004 +/- 0.003 | 0.005 +/- 0.003 | 6.71 +/- 4.69 | 5.74 +/- 5.09 |
| OCV_20 | 0.015 +/- 0.012 | 0.017 +/- 0.014 | 0.494 +/- 0.44 | 0.412 +/- 0.454 |
| R_30 | 0.003 +/- 0.002 | 0.004 +/- 0.002 | 5.19 +/- 3.51 | 4.31 +/- 3.85 |
| OCV_30 | 0.013 +/- 0.011 | 0.015 +/- 0.012 | 0.448 +/- 0.399 | 0.388 +/- 0.416 |
| R_40 | 0.002 +/- 0.002 | 0.003 +/- 0.002 | 3.81 +/- 2.63 | 3.14 +/- 2.9 |
| OCV_40 | 0.013 +/- 0.011 | 0.015 +/- 0.012 | 0.422 +/- 0.369 | 0.37 +/- 0.388 |
| R_50 | 0.002 +/- 0.002 | 0.002 +/- 0.002 | 3.81 +/- 3.02 | 3.22 +/- 3.28 |
| OCV_50 | 0.014 +/- 0.016 | 0.015 +/- 0.016 | 0.423 +/- 0.475 | 0.38 +/- 0.492 |
| R_60 | 0.002 +/- 0.001 | 0.002 +/- 0.002 | 4.01 +/- 3.07 | 3.41 +/- 3.29 |
| OCV_60 | 0.014 +/- 0.019 | 0.015 +/- 0.02 | 0.41 +/- 0.551 | 0.362 +/- 0.569 |
| R_70 | 0.002 +/- 0.002 | 0.002 +/- 0.002 | 5.01 +/- 4.58 | 4.28 +/- 4.84 |
| OCV_70 | 0.015 +/- 0.02 | 0.016 +/- 0.02 | 0.426 +/- 0.533 | 0.37 +/- 0.554 |
| R_80 | 0.002 +/- 0.002 | 0.003 +/- 0.002 | 4.33 +/- 3.24 | 3.52 +/- 3.38 |
| OCV_80 | 0.015 +/- 0.02 | 0.017 +/- 0.021 | 0.431 +/- 0.55 | 0.379 +/- 0.569 |
| R_90 | 0.002 +/- 0.002 | 0.002 +/- 0.002 | 2.99 +/- 2.33 | 2.4 +/- 2.4 |
| OCV_90 | 0.014 +/- 0.016 | 0.015 +/- 0.017 | 0.377 +/- 0.414 | 0.324 +/- 0.421 |

Table 3: Error metrics are provided for the CAMP test dataset SoPaH metrics.

| Metric | MAE | RMSE | MAPE | RMSPE |
|---|---|---|---|---|
| Q_d | 0.013 +/- 0.013 | 0.013 +/- 0.013 | 7.88 +/- 10.97 | 7.38 +/- 11.12 |
| E_d | 0.001 +/- 0.001 | 0.001 +/- 0.001 | 0.114 +/- 0.076 | 0.081 +/- 0.075 |
| Q_eff | 0.037 +/- 0.044 | 0.041 +/- 0.045 | 7.8 +/- 12.83 | 7.36 +/- 12.96 |
| E_eff | 0.005 +/- 0.004 | 0.006 +/- 0.004 | 0.571 +/- 0.475 | 0.538 +/- 0.491 |
| ohmic | 0.074 +/- 0.088 | 0.076 +/- 0.089 | 30.5 +/- 44.9 | 30.2 +/- 45.1 |

Figure 5a, c, and e show the mean, 70[th], and 95[th] percentiles of MAPE for the discharge capacity, discharge energy, and 50% SOC ECM resistance versus the number of predicted cycles for the fast-charge dataset. From these subfigures, 85% of capacity and energy predictions are within 3% error, and the same percentage of resistance predictions are within 10% error, all the way out to 2000 cycles. Prediction errors slightly increase versus cycles predicted up until high numbers of predicted cycles where data is scarce.

Figure 5b, d, and f show the mean, 70th, and 95th percentiles of MAPE for the discharge capacity, discharge energy, and 50% SOC ECM resistance, instead versus the start cycle for the prediction. Prediction errors for capacity and energy peak between start cycles 750 and 1000 at around 20% error for 85% of segments. The distribution of errors decreases dramatically after a 1250 start cycle as there are fewer relevant segments. Equivalent figures for the CAMP dataset are available in the SI.

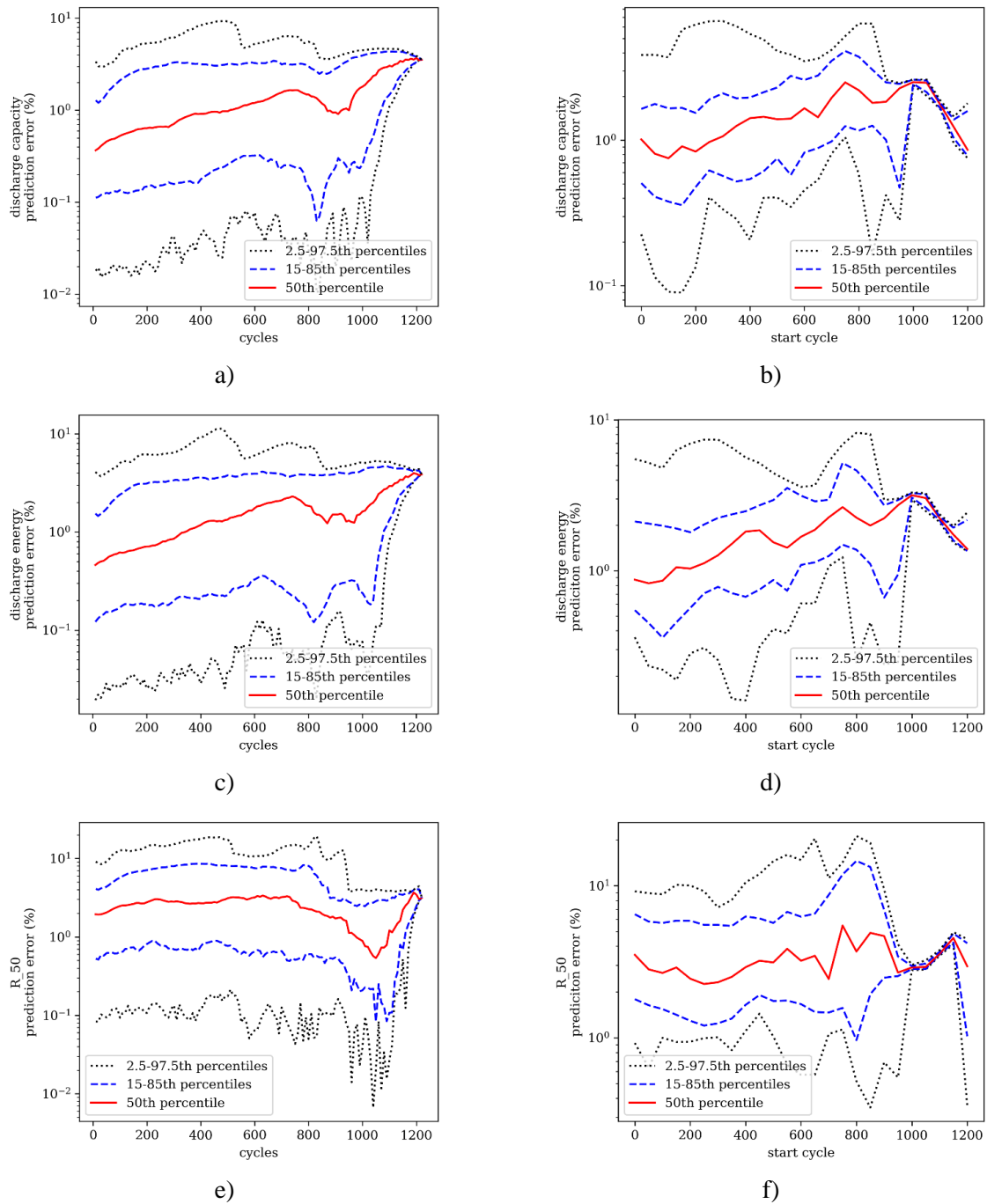

Figure 5: Prediction error versus cycle (a, c, and e) and average prediction error versus starting cycle (b, d, and f) for discharge capacity, discharge energy, and resistance at 50% SOC, respectively, for the fast-charge test dataset.

## 3.3 Prediction of EOL metrics

One of the most common performance metrics in the battery community presently is the number of cycles until a cell degrades to some percentage of initial capacity – often 80%. Figure 6 and Figure 7 plot predicted cycles versus experimental cycles until some EOL threshold (80% and 90% of the initial segment discharge capacity) was reached for each cycling segment in the fast-charging and CAMP datasets, respectively. To increase the number of valid segments, extrapolation was used to estimate EOL for segments that did not reach that threshold by the end of cycling. Extrapolation was performed using a linear model regressed on the last 10 or fewer cycles in each segment, with the weight on each cycle included in the regression linearly spaced from 0.01 to 1.0 from the first to the last cycle, respectively. *To avoid erroneous extrapolation, the regression slope was limited between -100 and -1000 times the initial capacity.* Figure 6 illustrates the life prediction accuracy for the fast-charging dataset. The prediction error is low, with mean absolute errors of approximately 19 and 33 cycles, for the 80% and 90% capacity EOL definitions, respectively. Prediction accuracy for the CAMP dataset when all cathode chemistries were included for training is illustrated in Figure 7, with MAEs of approximately 231 and 197 cycles, for the 80 and 90% capacity EOL definitions, respectively. Prediction error was roughly the same for cells corresponding to each cathode chemistry. This contrasts with our prior work with the CAMP dataset, where traditional ML models trained with all available cathode chemistries yielded different test errors for each chemistry (for example HE5050 had the highest test error in predicted cycle life). This reveals a difference in optimal training strategy between traditional ML and the DL approach employed here, though the reason for this difference is not obvious.

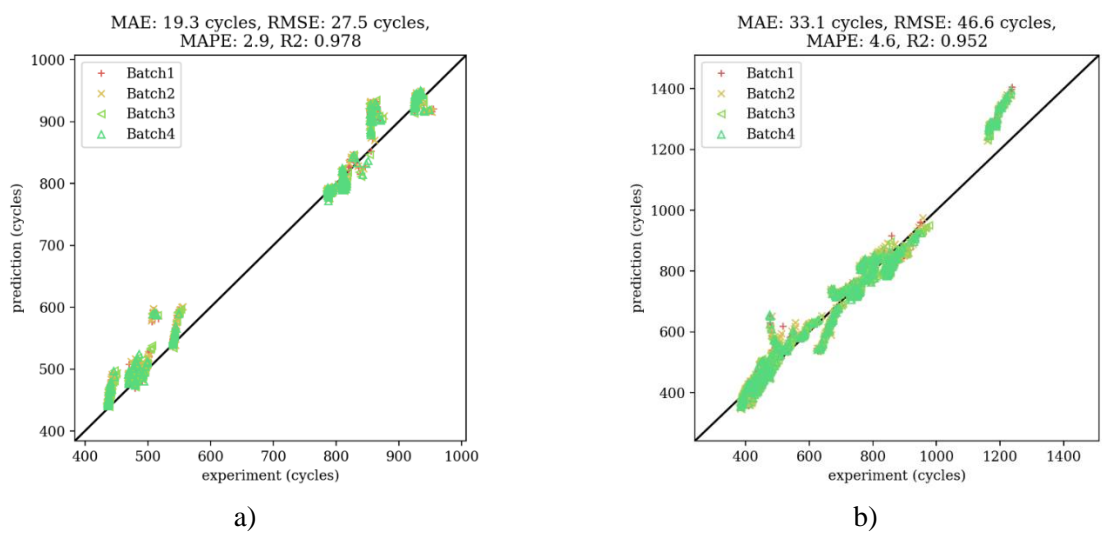

Figure 6: Experimental versus AI predicted number of cycles to a) 80%, and b) 90% of initial capacity for the fast-charge test dataset.

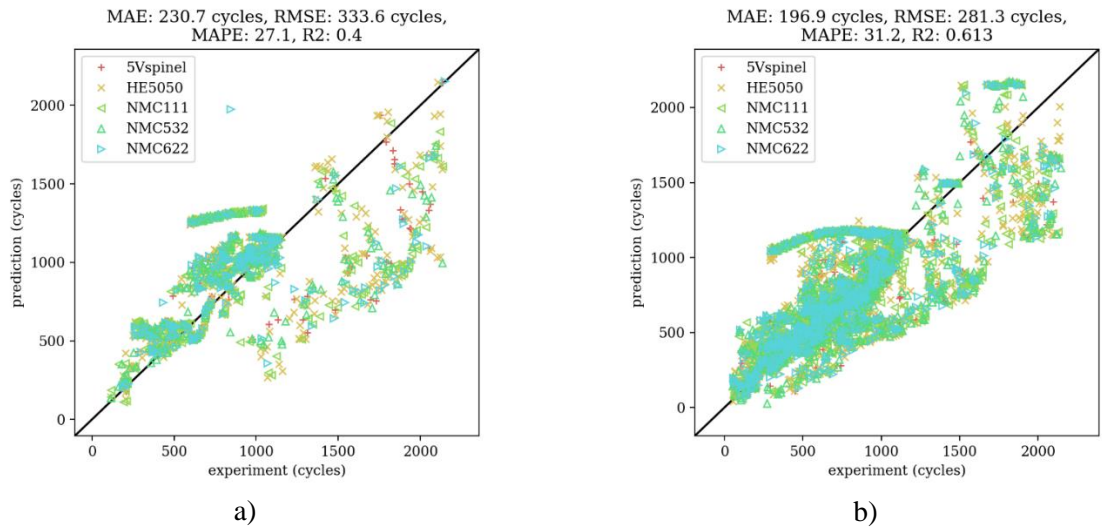

Figure 7: Experimental versus AI predicted number of cycles to a) 80%, and b) 90% of initial capacity for the CAMP test dataset.

Predictive performance was higher for some cathode chemistries when the training set was solely the same chemistry, as illustrated in Figure 8, Table 4, and Table 5. Predictive performance was especially high for the 5VSpinel and HE5050 chemistries, with MAEs of 76 and 102 cycles, respectively, when the EOL definition was 80% of the original discharge capacity. Predictive error for the NMC chemistries was considerably higher when trained separately than when the model was trained with all CAMP data simultaneously. In general, EOL prediction MAEs of approximately 100 cycles or fewer are consistent with our previous best results. Interestingly, for the 90% of original discharge capacity definitions, the experimental and predicted EOL cycles for the NMC datasets were highly correlated, exhibiting an offset from the true value. An additional model was trained using only cells with NMC cathodes but did not yield improved performance. In our prior work, traditional ML models trained only with 5VSpinel cells also had low test errors on 5VSpinel cells, but in contrast to this work, test error was low for the NMC cells for models trained only with NMC data.

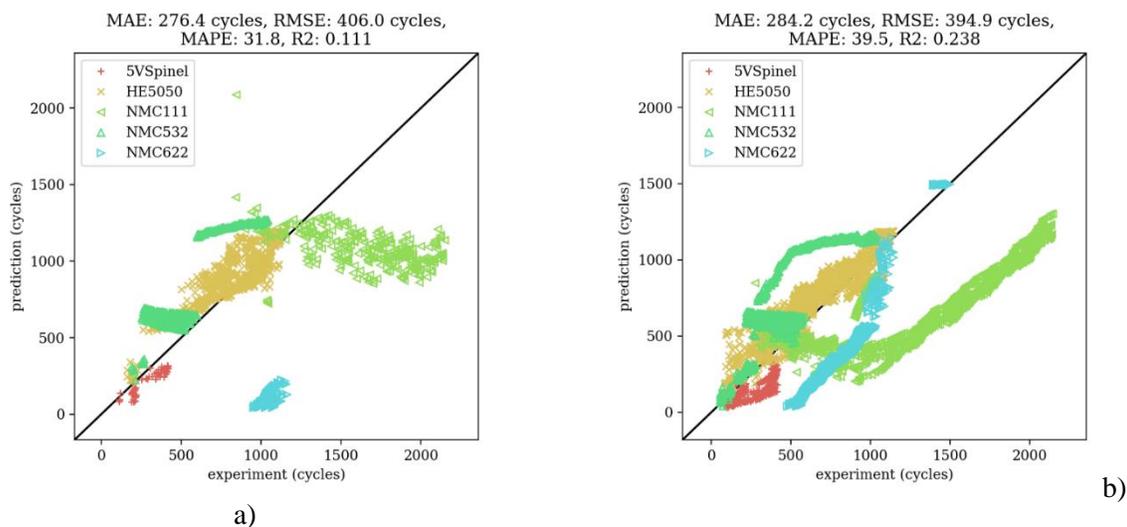

Figure 8: Experimental versus AI predicted number of cycles to a) 80%, and b) 90% of initial capacity for the HE5050, 5VSpinel, NMC111, NMC532, and NMC622 test datasets (each corresponding to the greater CAMP dataset).

To further explore these differences, an alternate training run was performed for the CAMP dataset where the NMC622 dataset was entirely excluded from the training and would therefore serve as a test of the model's predictive abilities for unseen cathode chemistries. The model was not highly predictive for this cathode chemistry (see Figure 10 in the SI) with MAEs of approximately 292 and 490 cycles, for the 80% and 90% capacity EOL definitions, respectively. Note that these models are not trained to specifically predict EOL; further work would be required to further incentivize the model to prioritize accurate prediction of this and related quantities.

Table 4: MAE and RMSE for the prediction of the number of cycles until 80% of the original capacity is reached. Errors are provided for each cathode fit all together or fit separately. The green shading indicates that the given method (fitting all cathodes together or fitting each separately) resulted in lower predictive error.

| cathode set | MAE fit-all (cycles) | MAE fit-indv. (cycles) | RMSE fit-all (cycles) | RMSE fit-indv. (cycles) |
|---|---|---|---|---|
| 5Vspinel | 244.3 | 76.2 | 340.7 | 85.4 |
| HE5050 | 228.7 | 102.5 | 336.0 | 126.1 |
| NMC111 | 231.4 | 598.1 | 326.7 | 685.6 |
| NMC532 | 230.6 | 289.5 | 329.3 | 325.9 |
| NMC622 | 226.1 | 942.6 | 339.2 | 943.5 |
| All | 230.7 | 276.4 | 333.6 | 406.0 |

Table 5: MAE and RMSE for the prediction of the number of cycles until 90% of the original capacity is reached. Errors are provided for each cathode fit all together or fit separately. The green shading indicates that the given method (fitting all cathodes together or fitting each separately) resulted in lower predictive error.

| cathode set | MAE fit-all (cycles) | MAE fit-indv. (cycles) | RMSE fit-all (cycles) | RMSE fit-indv. (cycles) |
|---|---|---|---|---|
| 5Vspinel | 193.6 | 115.1 | 279.1 | 133.3 |
| HE5050 | 186.7 | 101.9 | 268.3 | 123.0 |
| NMC111 | 213.3 | 656.4 | 300.7 | 715.4 |
| NMC532 | 209.2 | 243.6 | 296.4 | 298.0 |
| NMC622 | 184.2 | 371.5 | 263.2 | 400.6 |
| All | 196.9 | 284.2 | 281.3 | 394.9 |

# 4. Discussion

This work offers a generic approach to prognosis of multivariate battery performance and state of health trends using as few as a single cycle of current and voltage information as context. This capability can be leveraged to rapidly and comprehensively understand the effect of battery chemistry, design, and use scenario on degradation trends, which remains a critical need in the suite of energy storage innovation tools. Potential applications are broad and include, a) aiding the design of new battery chemistries and designs for target applications by anticipating dominant degradation modes, b) providing estimates of energy throughput and cost to de-risk grid-energy-storage investments, c) helping to optimize battery use throughout life to minimize degradation and maximize performance, and d) estimating the remaining value of batteries for secondary applications.

In most of the prior work, features were selected largely for their ability to inform the prediction of other quantities of interest, such as capacity curve knee points or EOL as defined by some reduction in capacity or resistance vs. the initial value. In the current work, however, what previously were features are now themselves also the predicted quantities of interest, and therefore the values of these features should ideally have physical or practical importance. Consequently, the value of the approach depends directly on the relevance and usefulness of these metrics. Unfortunately, in this work we employed a limited set of metrics due to limitations inherent in existing datasets. These initial results however can guide future work, where advanced techniques can be employed to provide access to a broad set of experimentally and computationally derived quantities.

Critically, reference performance tests (RPTs, or diagnostic cycling experiments performed on a regular interval) will provide a high-fidelity look into the battery. Common RPTs include capacity checks, evaluation of the round-trip efficiencies, and hybrid pulse power characterization tests that evaluate resistance as a function of the battery SOC. Furthermore, additional experimental techniques such as electrochemical impedance spectroscopy, and analytical techniques such as passive and active acoustic measurements, and in-situ X-ray spectroscopy and tomography [39], could provide an unprecedentedly detailed view of the evolution of the battery internal structure. Recent physics-based modeling efforts [25,40] may also provide insights into degradation mechanisms, modes, and their severity with battery use alongside detailed structural evolution information [41].

An interesting detail of the present work is the difference in the prediction error between the cells with different cathode chemistries. Accurate predictions were obtained for the LFP, HE5050, and 5VSpinel cathodes, but not for NMC111, NMC532, or NMC622. The discussion below proposes some explanations for this anomalous NMC predictive performance. Further work will be required to deconvolute prognosis accuracy from dataset details. Factors that may impact prognosis accuracy include the following:

a) <u>Dataset size</u>: The current datasets are relatively small in the total number of cells. Larger datasets may dramatically improve prognosis accuracy.

b) <u>Dataset diversity</u>: The fast-charge dataset was single chemistry (LFP) and dominated by a single degradation mechanism (lithium plating). In contrast, CAMP included six cathode chemistries and many degradation mechanisms. The chemical and mechanistic diversity of the CAMP dataset might have been a barrier to accurate prognosis. An alternate implication is that the NMC data "poisoned" prediction accuracy for the other cathode chemistries due to some inherent issues in the data. Some of the results support this theory, notably, that separate training for each cathode chemistry in the CAMP dataset resulted in lower prediction error for cells with HE5050 and 5VSpinel cathodes. Prior work with the same dataset exhibited lower prediction error with NMC cathodes; however, in that work, traditional ML techniques were used and the goal was predicting end of life, not to predict the entire degradation trend as in this work.

c) <u>Data collection</u>: The fast-charge data collection was rarely interrupted, and no secondary testing was performed. In contrast, CAMP cycling was regularly interrupted by a variety of RPTs at intervals that varied between cells. These interruptions may have negatively impacted prognosis accuracy.

d) <u>Data processing</u>: It is presently unclear how and why data smoothing affects prognosis accuracy. While smoothing may be necessary to reduce noise and outliers in the data, it may corrupt subtle correlations between and within the multiple time-series, such as those related to measurement noise itself.

Overall, the multivariate transformer approach predicted performance and degradation trends versus cycling for a variety of cathode chemistries and two use scenarios, a) standard low C-rate charge and discharge cycling and b) diverse fast-charging scenarios with standard high-rate discharge. Prognosis error either stayed constant or slightly increased as the forecast horizon became larger. In contrast, average prognosis error was not highly correlated with cell history. In other words, the algorithm predicted degradation trends equally well for fresh cells as for cells that had already seen up to one thousand cycles.

# 5. Conclusions

In this work, we present an advancement to the state of the art of lithium-ion battery prognosis which not only offers projections across the entire cycling experiment but does this using a single preliminary cycle as context and for multiple dimensions of performance. Further, this is possible not only with capacities, but also with round-trip efficiencies, open circuit voltage values, and resistances across two datasets and seven cathode chemistries. Predictions were accurate with a mean absolute percent error (MAPE) of 2.7% in the prediction of the number of cycles to 80% of the original capacity for a lithium-iron-phosphate fast-charging dataset. Although the composition of the test set differs, the previous best-known prediction was 7% MAPE [17].

This advancement was made possible by a specialized version of the deep learning transformer architecture, now famous as the basis for tremendous achievements in natural language processing, that uses the attention mechanism to identify key relationships between elements both within and across input and output sequences. Although the network has no knowledge of the physics at play, its ability to learn relationships between important quantities results in remarkably accurate predictions of complex phenomena – an observation of deep learning performance that has been mirrored across the sciences.

The approach is remarkable not only in its capability to predict the evolution of multiple electrochemical parameters simultaneously, but also by its promise to integrate non-electrochemical values such as materials properties. In future work, we will extend prognosis to structural characteristics including electrode morphology, solid electrolyte interface thickness and coverage, and electrolyte health. These quantities can be measured or estimated via non-destructive means (e.g., electrochemical impedance spectroscopies, acoustic interrogations, in-situ x-ray, and physics-based modeling). Applying the techniques presented in this work to a comprehensive set of performance and health metrics may unlock a previously impossible prognosis ability – to "see" inside the battery at any future point in time, which is critical to hastening the pace of energy storage innovations that support the development of a fully decarbonized energy sector.

# Acknowledgements

The authors acknowledge financial support from Laboratory Directed Research and Development (LDRD) funding from Argonne National Laboratory, provided by the Director, Office of Science, of the U.S. Department of Energy under Contract No. DE-AC02-06CH11357 and from the U.S. Department of Energy Office of Electricity. We also gratefully acknowledge the computing resources provided by the Laboratory Computing Resource Center on the Bebop and Swing supercomputers at Argonne National Laboratory. The electrodes, pouch cells, and 18650 cells were produced at the U.S. Department of Energy's (DOE) CAMP (Cell Analysis, Modeling and Prototyping) Facility, Argonne National Laboratory. The CAMP Facility is fully supported by the DOE Vehicle Technologies Office (VTO). N.H. Paulson acknowledges Michael Shuffett for discussions on transformers that greatly aided this work.

# Data and Code Availability

Processed training data is available on the Materials Data Facility [42]. All codes required to reproduce these findings are available on GitHub [34].

# Declaration of Interests

The authors declare no competing interests.

# CRediT Author Statement

**Noah H. Paulson**: Conceptualization, Methodology, Software, Writing – original draft, Writing – review & editing, Visualization, Project administration, Funding acquisition, **Joseph J. Kubal**: Methodology, Data curation, Writing – review & editing, **Susan J. Babinec**: Conceptualization, Writing – review & editing, Funding acquisition

# Supplementary Information

The following is the command used to run the spacetimeformer model including critical hyperparameters:

```
>>> train.py spacetimeformer battery_CAMP --run_name
spatiotemporal_battery_CAMP --d_model 100 --d_ff 200 --enc_layers
4 --dec_layers 4 --gpus 0 --batch_size 4 --start_token_len 4 --
n_heads 4 --context_points 1 --skip_context 1 --skip_target 10 --
early_stopping --wandb --plot
```

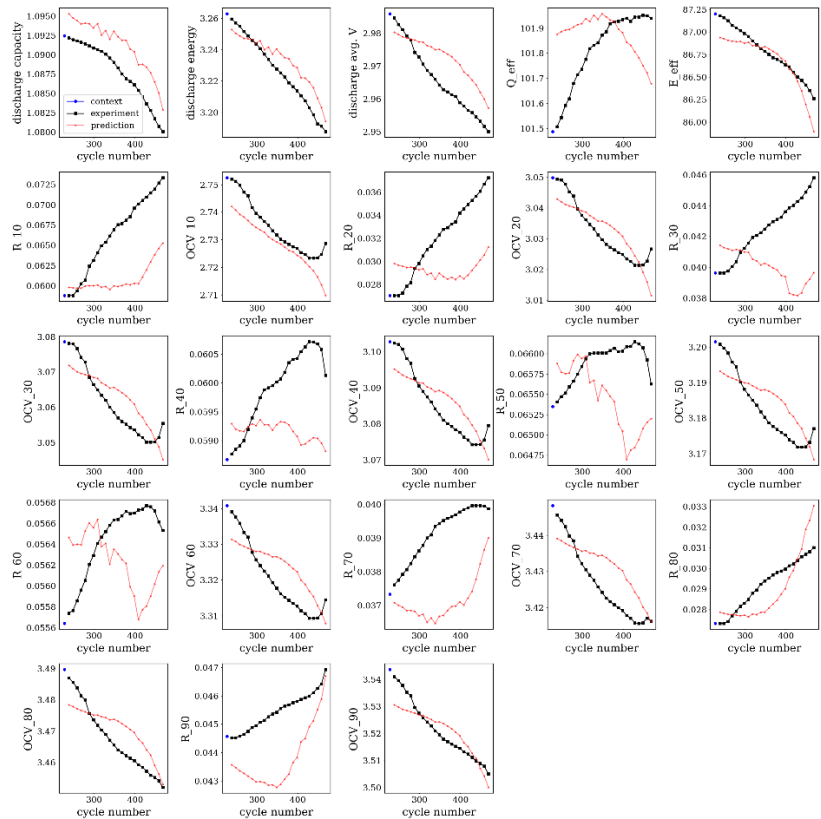

Figure 9: Prediction versus experiment for cell CH37 of batch 1 between cycles 238 and 469 across all SOPAH metrics showing poor predictive ability.

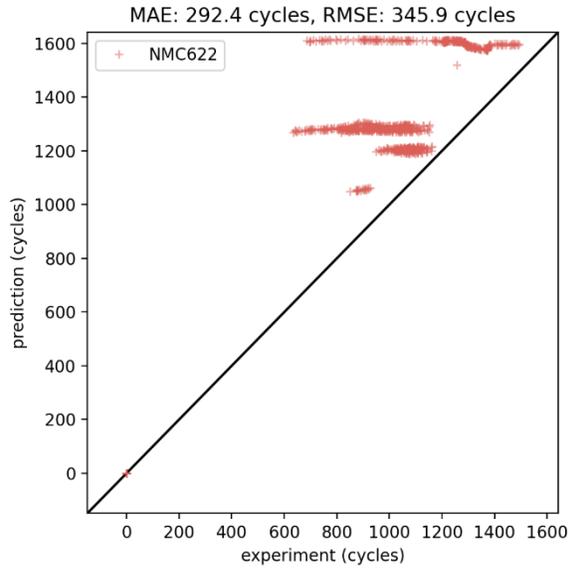 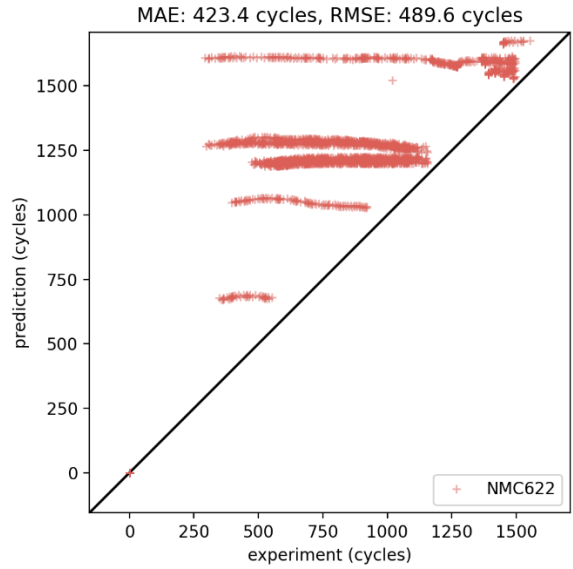

Figure 10: Experimental versus AI predicted number of cycles to a) 80%, and b) 90% of initial capacity for the CAMP new cathode dataset.

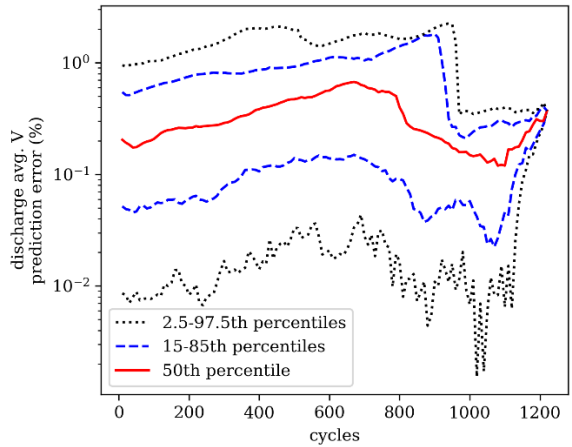 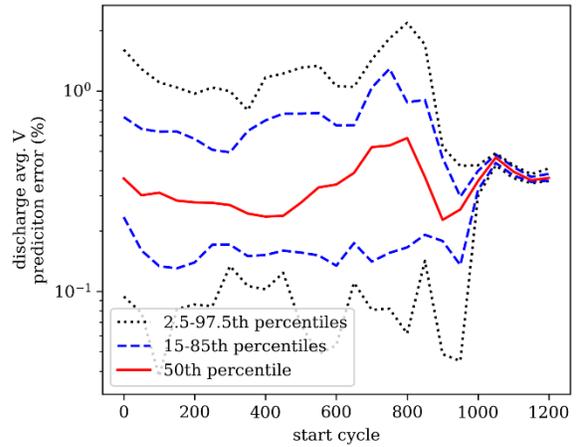

Figure 11: a) Prediction error versus cycle and b) average test prediction error versus starting cycle for the average discharge voltage in the fast-charge test dataset.

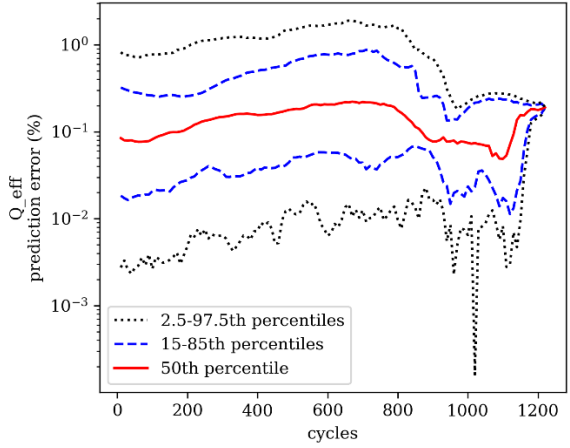
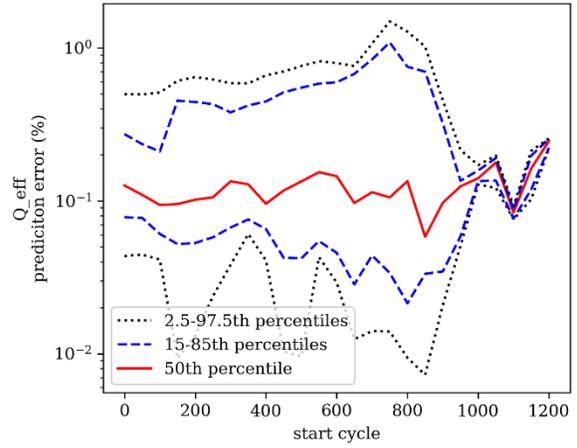

Figure 12: a) Prediction error versus cycle and b) average test prediction error versus starting cycle for the Coulombic efficiency in the fast-charge test dataset.

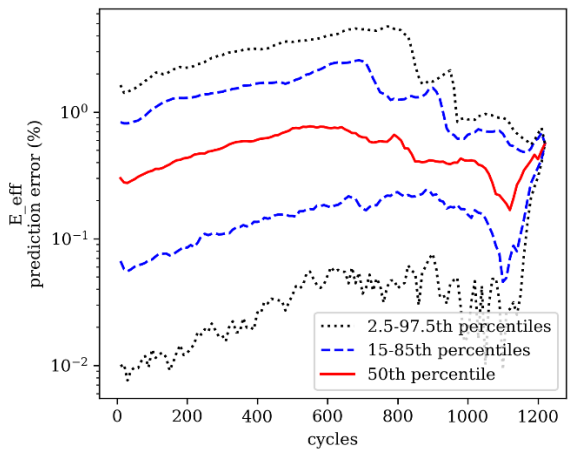
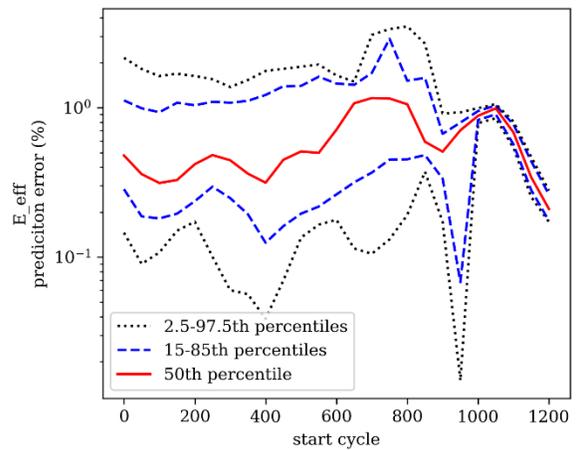

Figure 13: a) Prediction error versus cycle and b) average test prediction error versus starting cycle for the energy efficiency in the fast-charge test dataset.

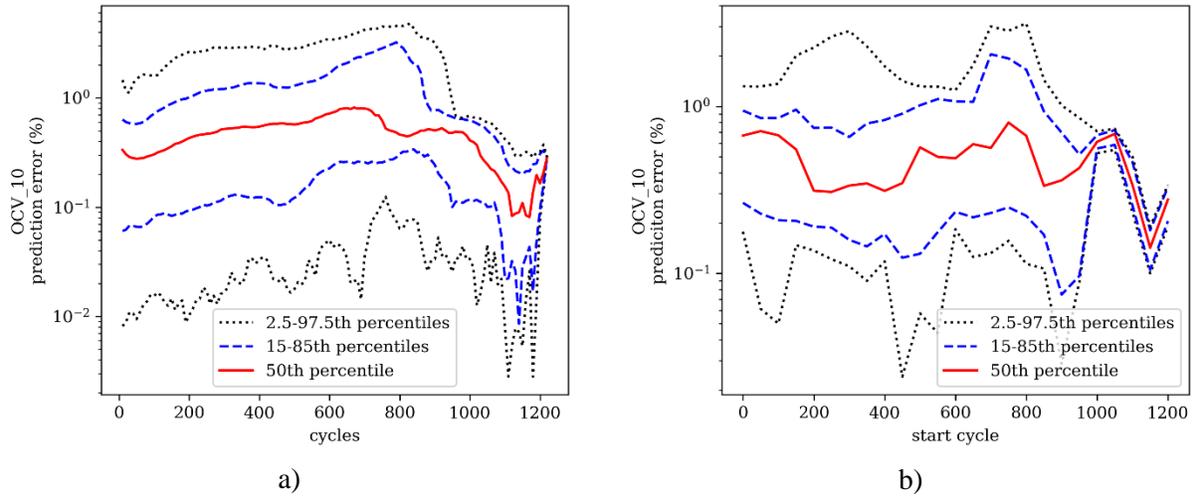

Figure 14: a) Prediction error versus cycle and b) average test prediction error versus starting cycle for the ECM-fit OCV at 10%SOC in the fast-charge test dataset.

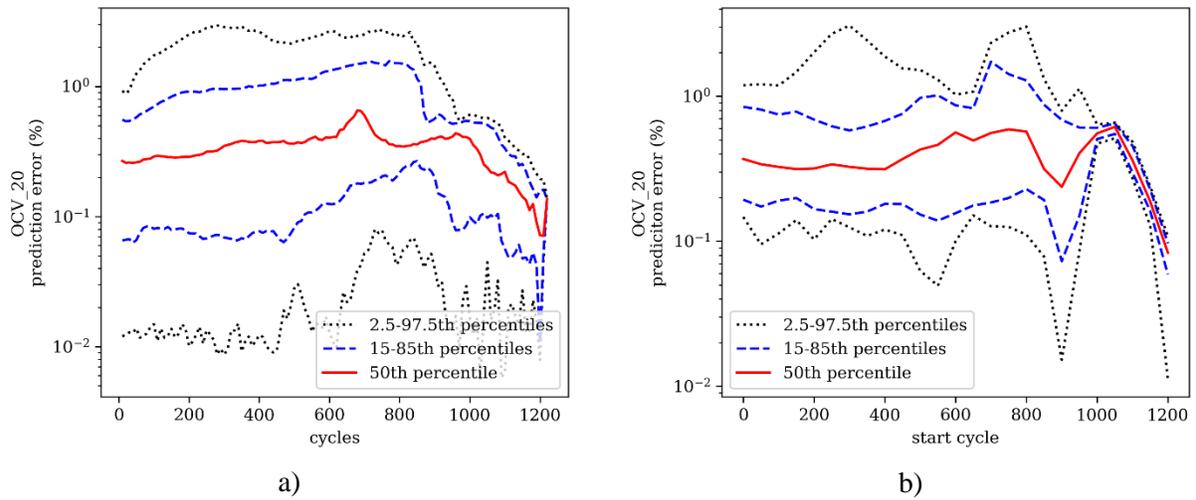

Figure 15: a) Prediction error versus cycle and b) average test prediction error versus starting cycle for the ECM-fit OCV at 20%SOC in the fast-charge test dataset.

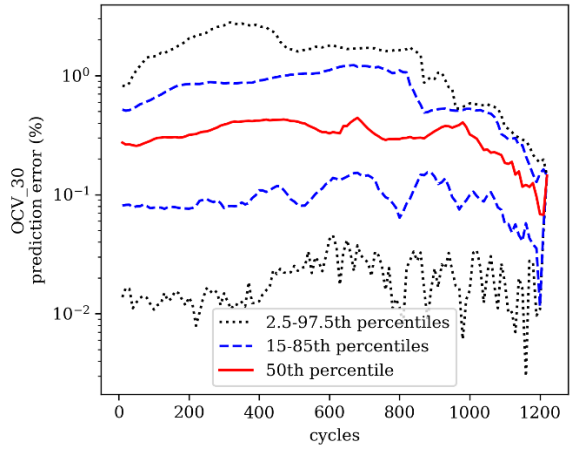
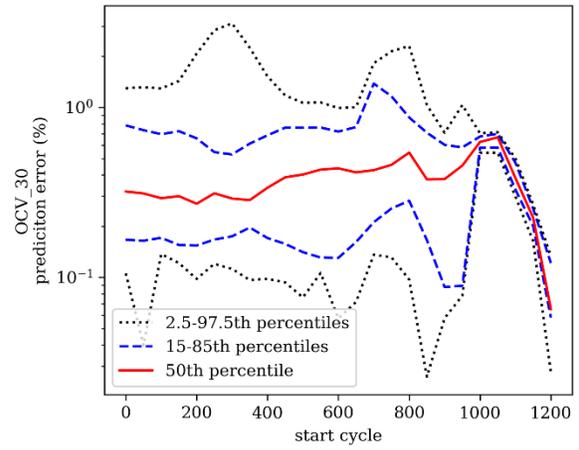

Figure 16: a) Prediction error versus cycle and b) average test prediction error versus starting cycle for the ECM-fit OCV at 30%SOC in the fast-charge test dataset.

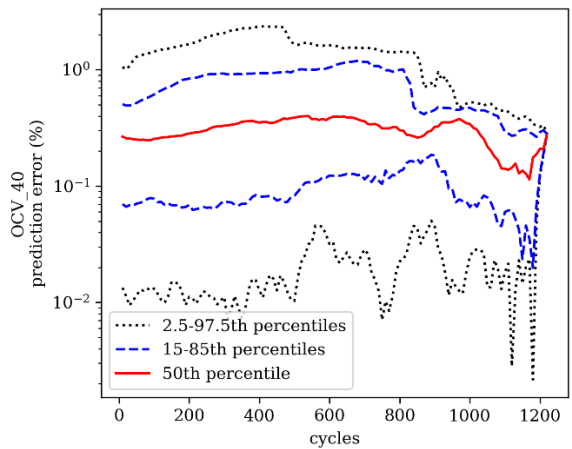
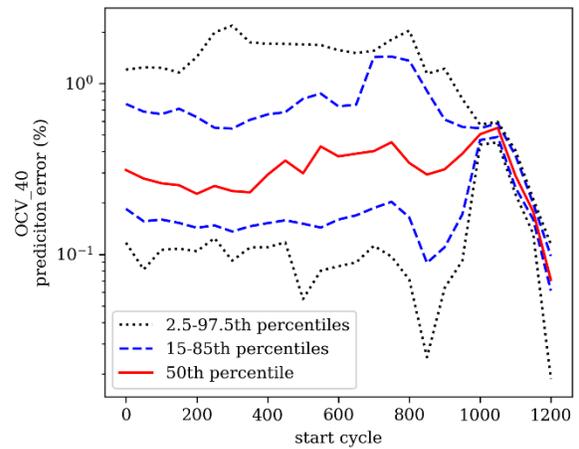

Figure 17: a) Prediction error versus cycle and b) average test prediction error versus starting cycle for the ECM-fit OCV at 40%SOC in the fast-charge test dataset.

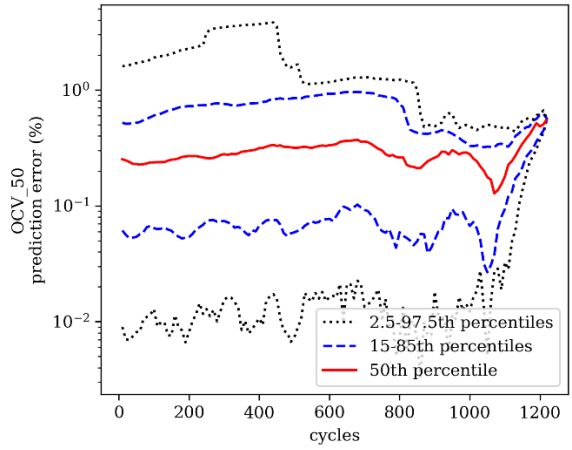 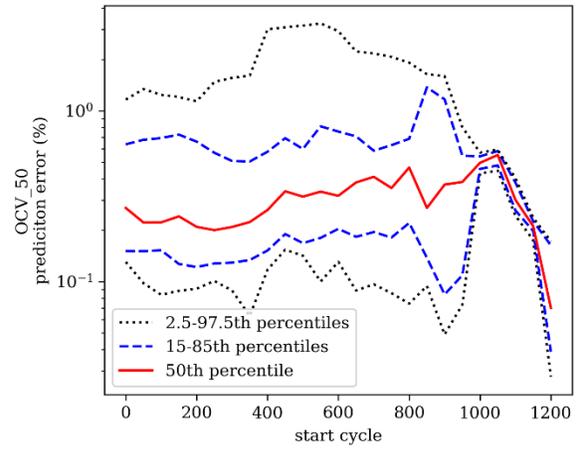

Figure 18: a) Prediction error versus cycle and b) average test prediction error versus starting cycle for the ECM-fit OCV at 50%SOC in the fast-charge test dataset.

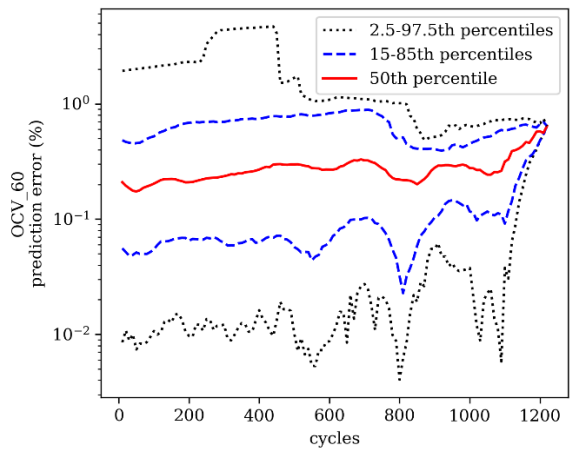 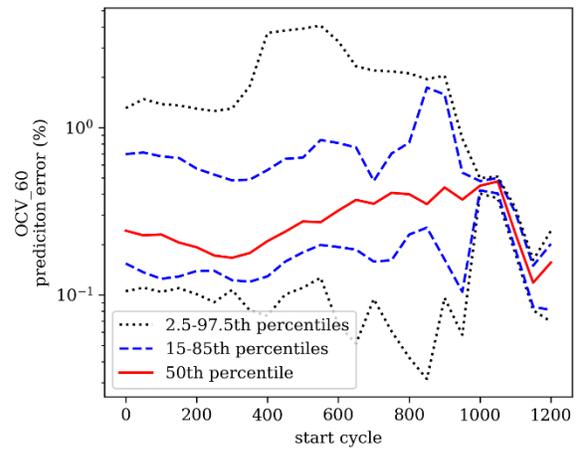

Figure 19: a) Prediction error versus cycle and b) average test prediction error versus starting cycle for the ECM-fit OCV at 60%SOC in the fast-charge test dataset.

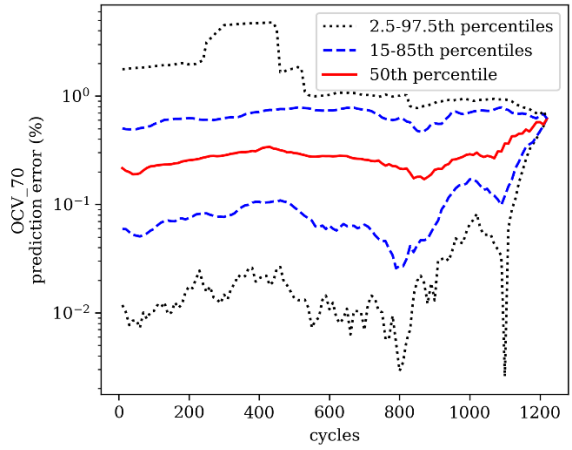
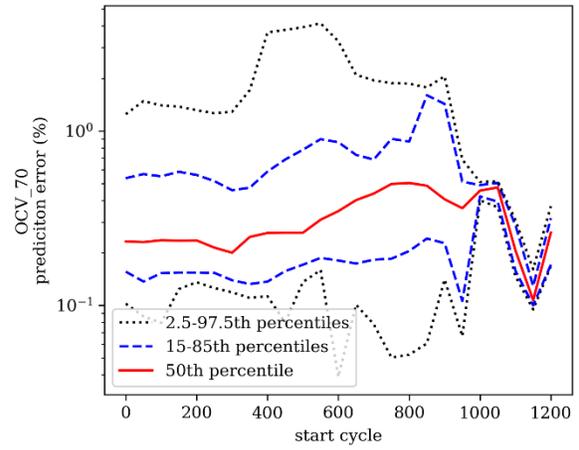

Figure 20: a) Prediction error versus cycle and b) average test prediction error versus starting cycle for the ECM-fit OCV at 70%SOC in the fast-charge test dataset.

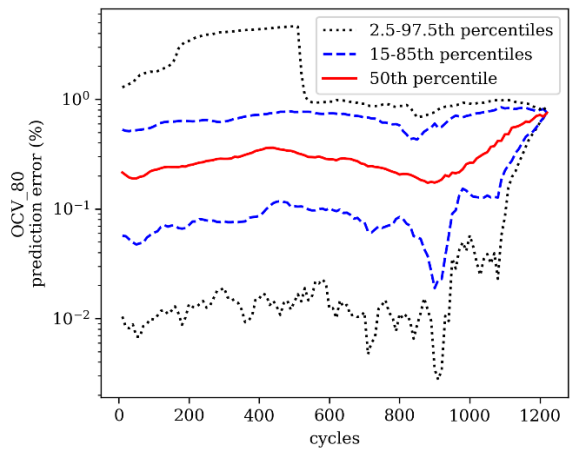
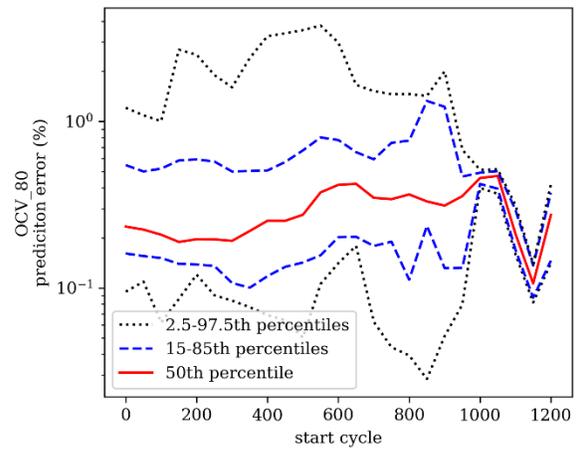

Figure 21: a) Prediction error versus cycle and b) average test prediction error versus starting cycle for the ECM-fit OCV at 80%SOC in the fast-charge test dataset.

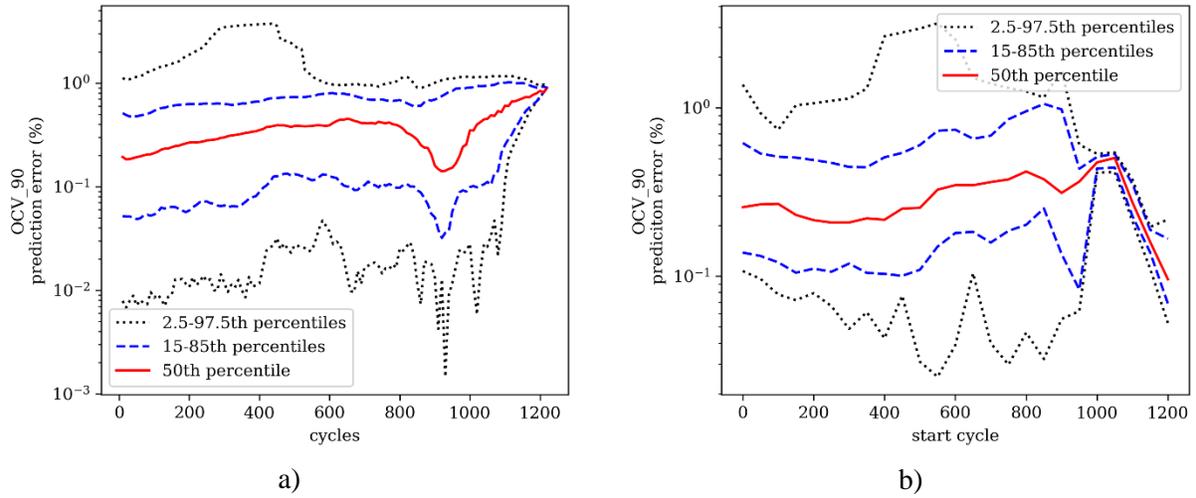

Figure 22: a) Prediction error versus cycle and b) average test prediction error versus starting cycle for the ECM-fit OCV at 90%SOC in the fast-charge test dataset.

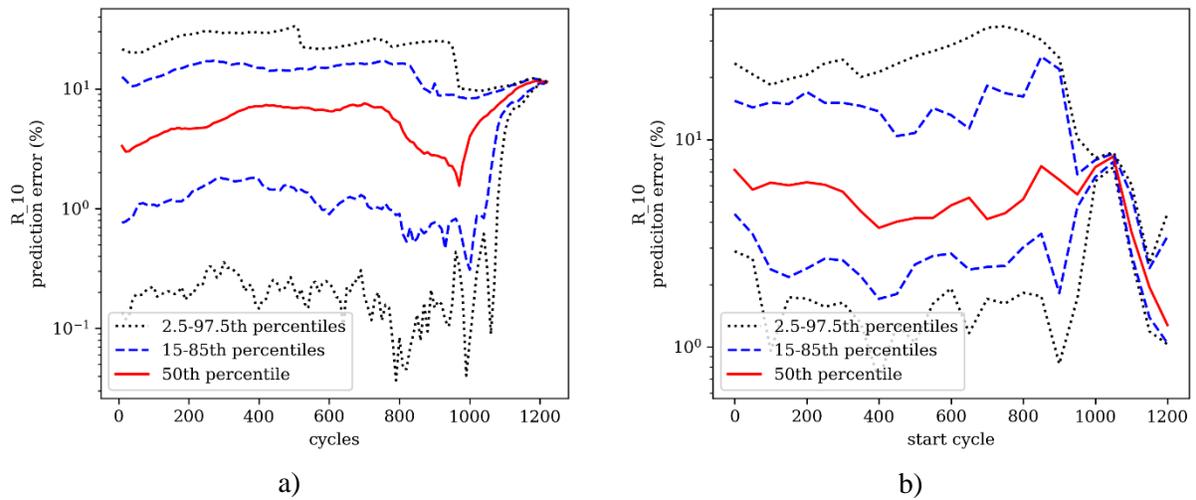

Figure 23: a) Prediction error versus cycle and b) average test prediction error versus starting cycle for the ECM-fit resistance at 10%SOC in the fast-charge test dataset.

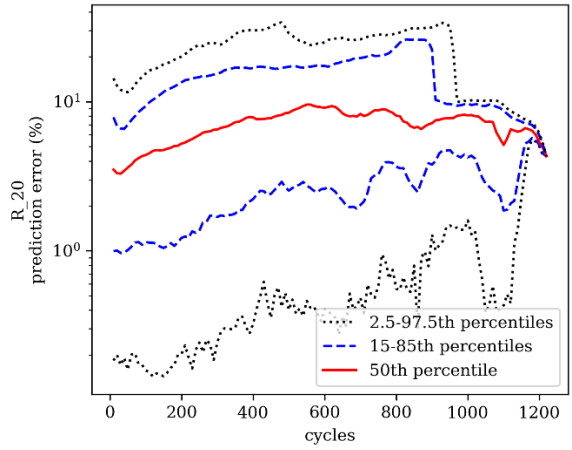
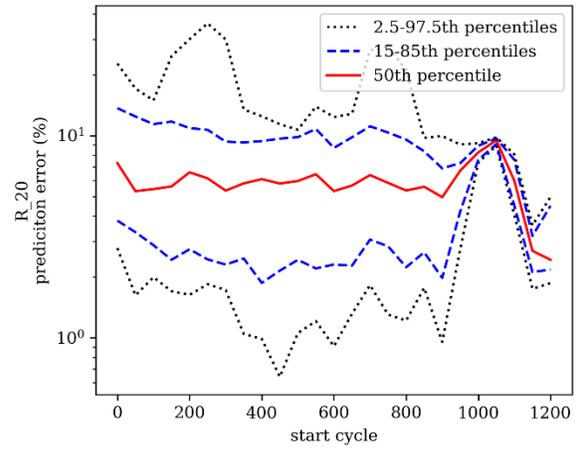

Figure 24: a) Prediction error versus cycle and b) average test prediction error versus starting cycle for the ECM-fit resistance at 20%SOC in the fast-charge test dataset.

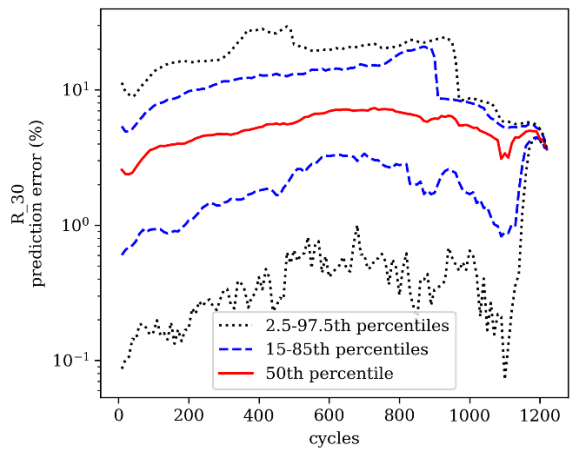
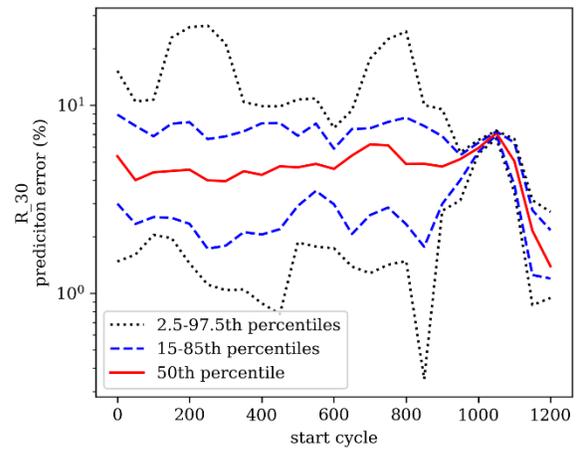

Figure 25: a) Prediction error versus cycle and b) average test prediction error versus starting cycle for the ECM-fit resistance at 30%SOC in the fast-charge test dataset.

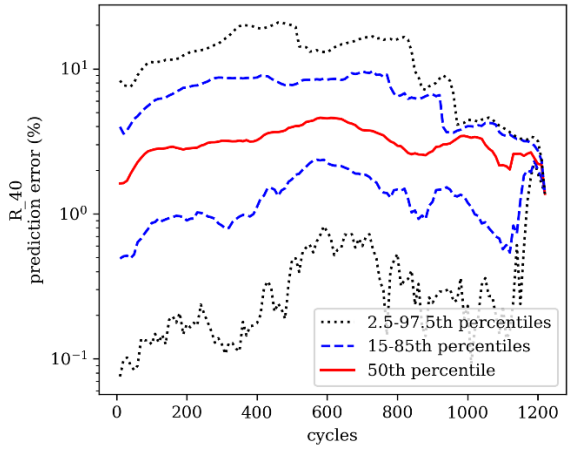
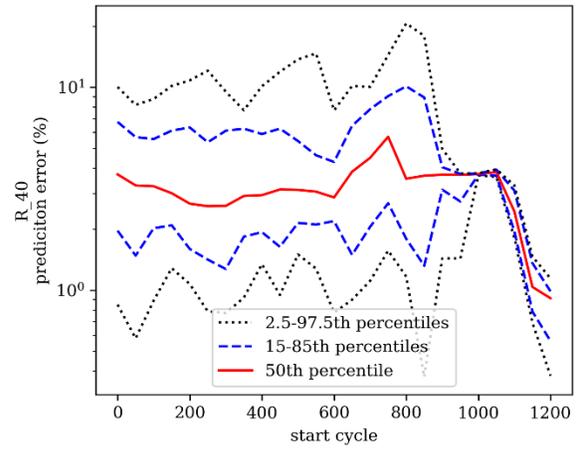

Figure 26: a) Prediction error versus cycle and b) average test prediction error versus starting cycle for the ECM-fit resistance at 40%SOC in the fast-charge test dataset.

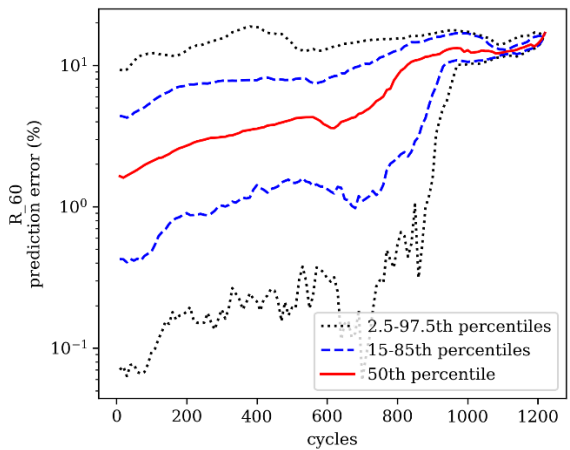
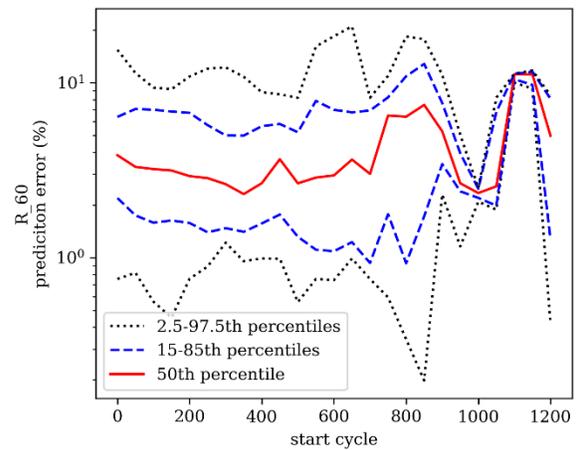

Figure 27: a) Prediction error versus cycle and b) average test prediction error versus starting cycle for the ECM-fit resistance at 60%SOC in the fast-charge test dataset.

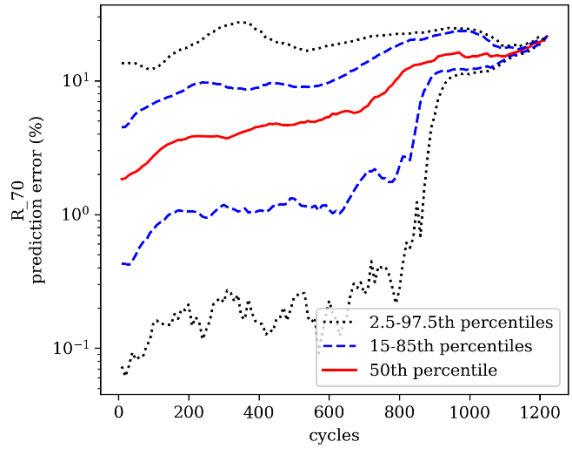 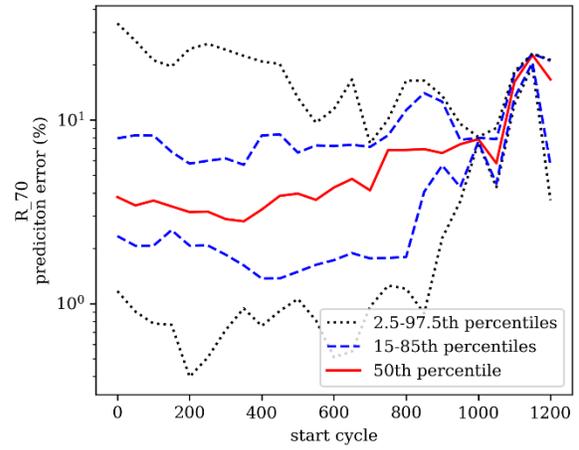

a) b)

Figure 28: a) Prediction error versus cycle and b) average test prediction error versus starting cycle for the ECM-fit resistance at 70%SOC in the fast-charge test dataset.

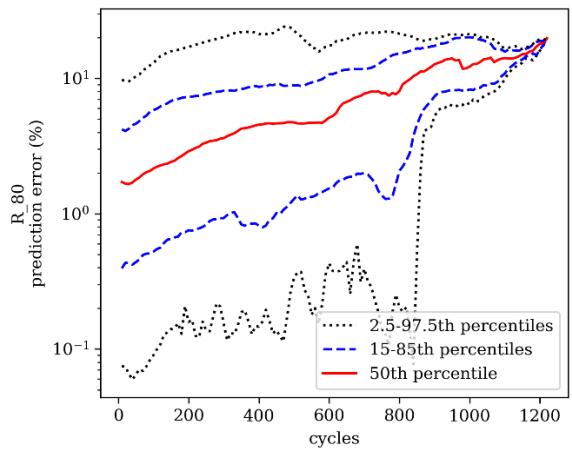 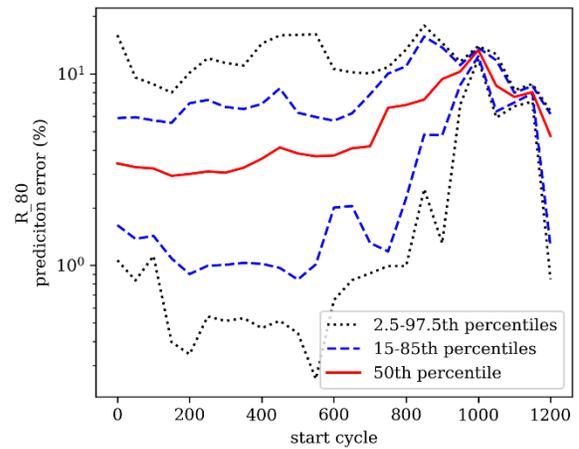

a) b)

Figure 29: a) Prediction error versus cycle and b) average test prediction error versus starting cycle for the ECM-fit resistance at 80%SOC in the fast-charge test dataset.

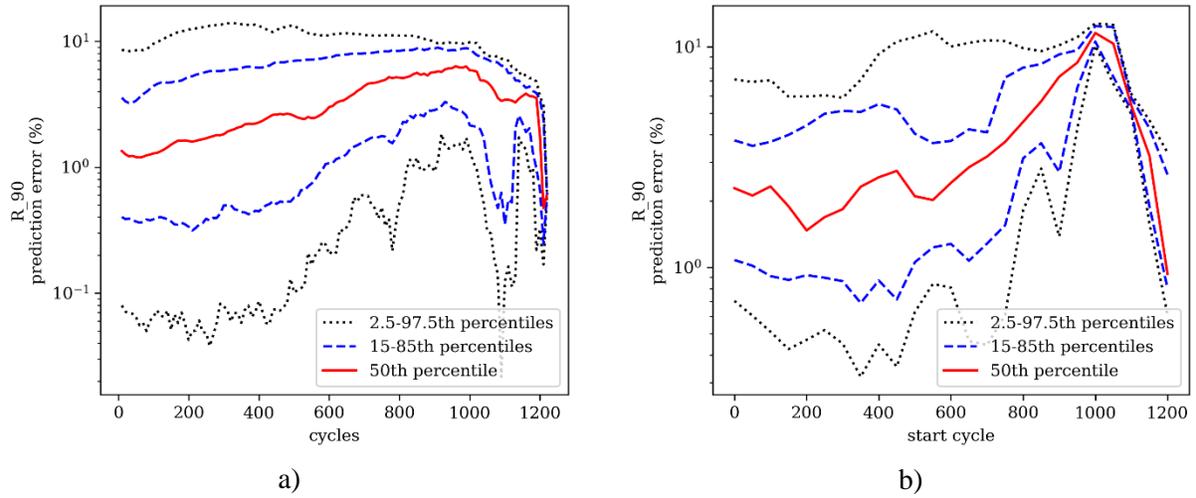

Figure 30: a) Prediction error versus cycle and b) average test prediction error versus starting cycle for the ECM-fit resistance at 90%SOC in the fast-charge test dataset.

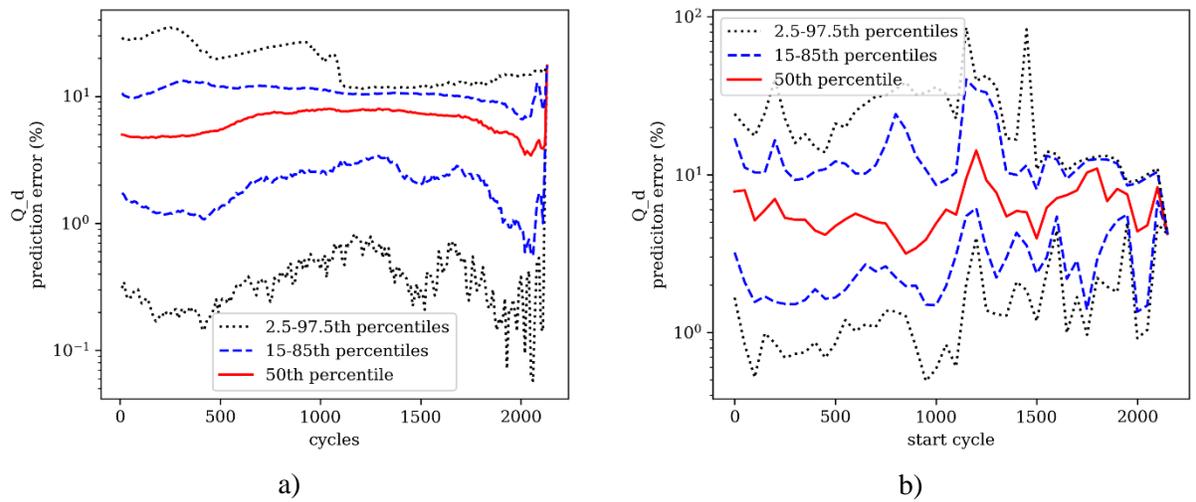

Figure 31: a) Prediction error versus cycle and b) average test prediction error versus starting cycle for the discharge capacity in the CAMP test dataset.

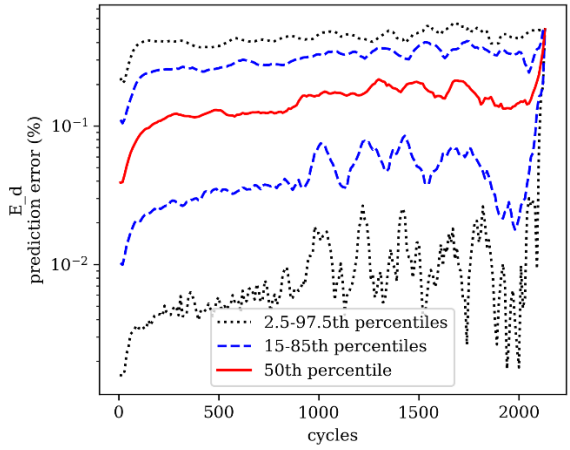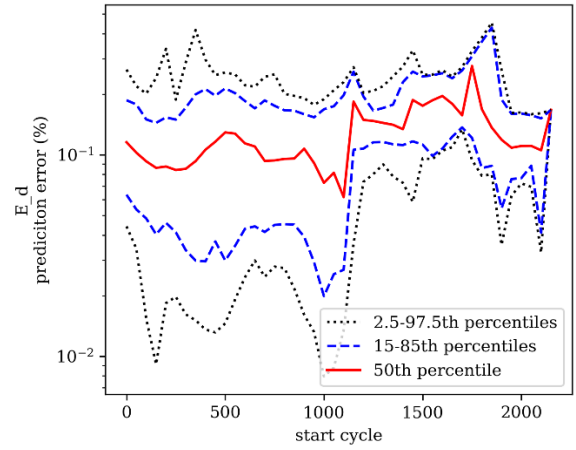

Figure 32: a) Prediction error versus cycle and b) average test prediction error versus starting cycle for the discharge energy in the CAMP test dataset.

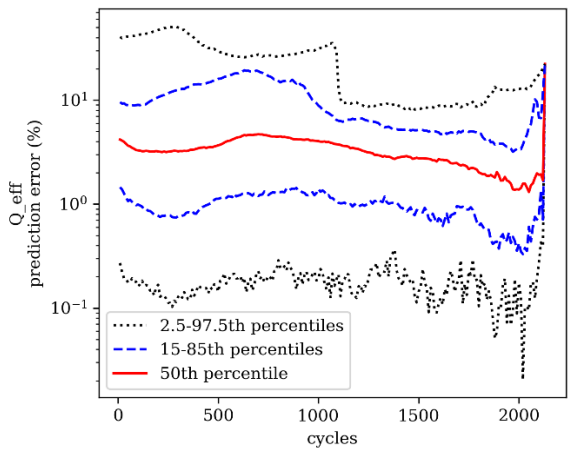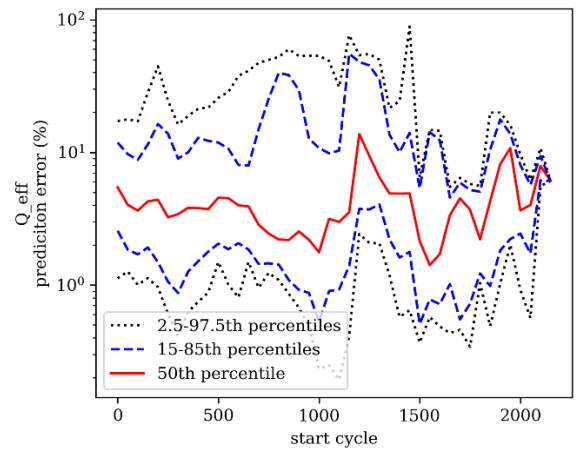

Figure 33: a) Prediction error versus cycle and b) average test prediction error versus starting cycle for the coulombic efficiency in the CAMP test dataset.

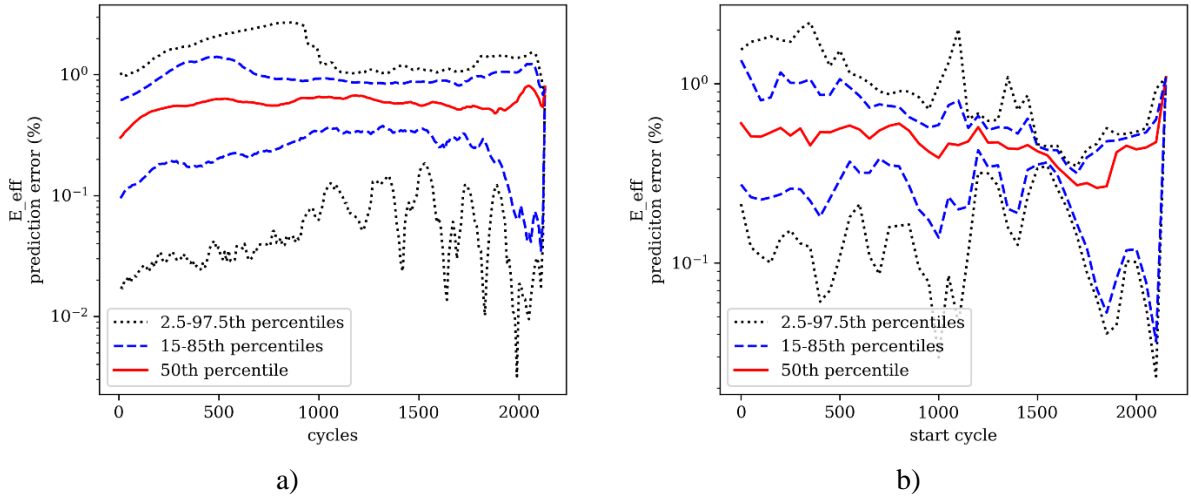

Figure 34: a) Prediction error versus cycle and b) average test prediction error versus starting cycle for the round-trip energy efficiency in the CAMP test dataset.

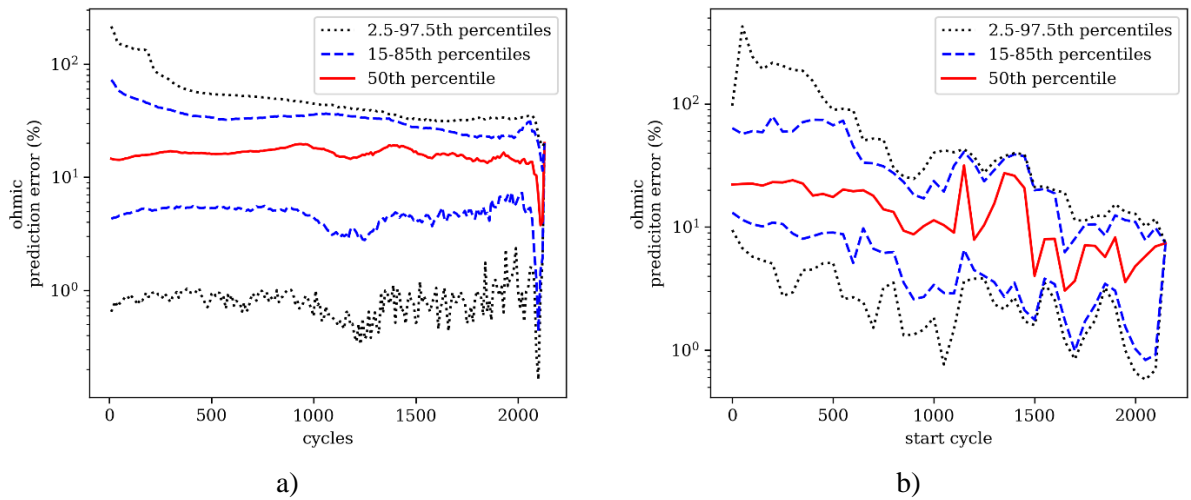

Figure 35: a) Prediction error versus cycle and b) average test prediction error versus starting cycle for the ohmic resistance in the CAMP test dataset.